\documentclass[journal]{IEEEtran}
\usepackage{cite}
\usepackage{amsmath,amssymb,amsfonts}
\usepackage{algorithmic}
\usepackage{graphicx}
\usepackage{textcomp}
\usepackage{hyperref}
\usepackage{booktabs}
\usepackage{threeparttable}
\usepackage{graphicx} 
\usepackage{bm}
\usepackage{eurosym}
\usepackage{adjustbox}

\begin{document}

\title{A Systematic Review of Digital Twin-Driven Predictive Maintenance in Industrial Engineering: Taxonomy, Architectural Elements, and Future Research Directions}

\author{Leila Ismail*,~\IEEEmembership{Member,~IEEE,}
        Abdelmoneim Abdelmoti,
        Arkaprabha Basu,
        Aymen Dia Eddine Berini,
        and Mohammad Naouss
\thanks{(* Corresponding author: Leila Ismail.)}%
\thanks{Dr. Leila Ismail is with Intelligent Distributed Computing and Systems (INDUCE) Lab, Department of Computer Science and Software Engineering, College of Information Technology, United Arab Emirates University, Al-Ain, United Arab Emirates, Emirates Center for Mobility Research, United Arab Emirates University, Al-Ain, United Arab Emirates, (leila@uaeu.ac.ae)}%
\thanks{Abdelmoneim Abdelmoti was with Department of Architectural Engineering, College of Engineering, United Arab Emirates University, Al-Ain, United Arab Emirates (Work contributed while student at United Arab Emirates University)}%
\thanks{Arkaprabha Basu is with Intelligent Distributed Computing and Systems (INDUCE) Lab, Department of Computer Science and Software Engineering, College of Information Technology, United Arab Emirates University, Al-Ain, United Arab Emirates, Emirates Center for Mobility Research, United Arab Emirates University, Al-Ain, United Arab Emirates}
\thanks{Aymen Dia Eddine Berini is with Department of Information Systems and Security, College of Information Technology, United Arab Emirates University, Al-Ain, United Arab Emirates}
\thanks{Mohammad Naouss is with Department of Computer  and Network Engineering, United Arab Emirates University, Al-Ain, United Arab Emirates}
}


\maketitle

\begin{abstract}
With the increasing complexity of industrial systems, there is a pressing need for predictive maintenance to avoid costly downtime and disastrous outcomes that could be life-threatening in certain domains. With the growing popularity of the Internet of Things, Artificial Intelligence, machine learning, and real-time big data analytics, there is a unique opportunity for efficient predictive maintenance to forecast equipment failures for real-time intervention and optimize maintenance actions, as traditional reactive and preventive maintenance practices are often inadequate to meet the requirements for the industry to provide quality-of-services of operations. Central to this evolution is digital twin technology, an adaptive virtual replica that continuously monitors and integrates sensor data to simulate and improve asset performance. Despite remarkable progress in digital twin implementations, such as considering DT in predictive maintenance for industrial engineering. This paper aims to address this void. We perform a retrospective analysis of the temporal evolution of the digital twin in predictive maintenance for industrial engineering to capture the applications, middleware, and technological requirements that led to the development of the digital twin from its inception to the AI-enabled digital twin and its self-learning models. We provide a layered architecture of the digital twin technology, as well as a taxonomy of the technology-enabled industrial engineering applications systems, middleware, and the used Artificial Intelligence algorithms. We provide insights into these systems for the realization of a trustworthy and efficient smart digital-twin industrial engineering ecosystem. We discuss future research directions in digital twin for predictive maintenance in industrial engineering.
\end{abstract}


\section{Introduction}
\label{sec:Introduction}
In the era of industry $5.0$, and the rise of smart city applications, enabled by Industrial Internet of Things (IIoT), big data analytics, Artificial Intelligence (AI), distributed federated learning, edge and cloud computing, smart factories \cite{statistaTopicSmart} emerge to replace traditional ones. Digital Twin (DT) is a virtual replica of a physical component that mirrors real-time operations but predicts future conditions and operations.

Nowadays, industrial systems, whether in manufacturing plants, energy facilities, or transportation networks, are engineered with an expected life expectancy of about 10 years \cite{aivaliotis2021degradation}. However, exposure to constant degrading influences such as overloading, extreme weather, and material deterioration gradually degrades these assets, leading to unexpected failures and interruptions. It is reported that industrial engineering components failures increased by 40\% between 2010 and 2018 \cite{corchado2019failure}. In addition, a single day of unplanned downtime can cost between $100,000$ and $200,000$ \euro \cite{mathur2001reasoning}, and maintenance expenses can consume up to $60$ - $70$\% of the operational life of an asset cost \cite{aivaliotis2021degradation}. 

Maintenance strategies evolved in 3 directions: 1) reactive maintenance, as repairs are made after failures leading to prolonged downtime and unpredictable expenses \cite{mobley2002introduction, alnajar2003selecting, moubray1997reliability}, 2) preventive, which constitutes scheduled maintenance at fixed intervals regardless of the actual condition of the equipment. Although this strategy reduced sudden breakdowns, it often resulted in premature replacement of parts and inefficient utilization of resources \cite{jardine2006maintenance, dhillon2002engineering, ebeling1997introduction, wireman2007preventive}, to 3) predictive maintenance, leveraging continuous monitoring through real-time sensor data and advanced analytics. By predicting when a failure is likely to occur, PdM allows targeted interventions that minimize downtime and optimize maintenance schedules \cite{rojek2023artificial, mobley2002introduction}. This procedure helps to determine \textit{the failure before it happens} and integrate reflex precautionary measures to tackle the situation.

The integration of DT technology with PdM is paramount in the industry $5.0$ revolution and the design of smart city \cite{ismail2018information}. A DT enables a digital replica to update itself in lockstep with its physical counterpart, a system that not only reflects current conditions but also simulates future scenarios using machine learning and deep learning, and large language models. The DT concept emerged from early simulation in aerospace and matured into a sensor-driven digital model. \cite{rojek2023artificial,aivaliotis2021degradation}.

\begin{figure}[ht]
    \centering
    \includegraphics[width=\columnwidth]{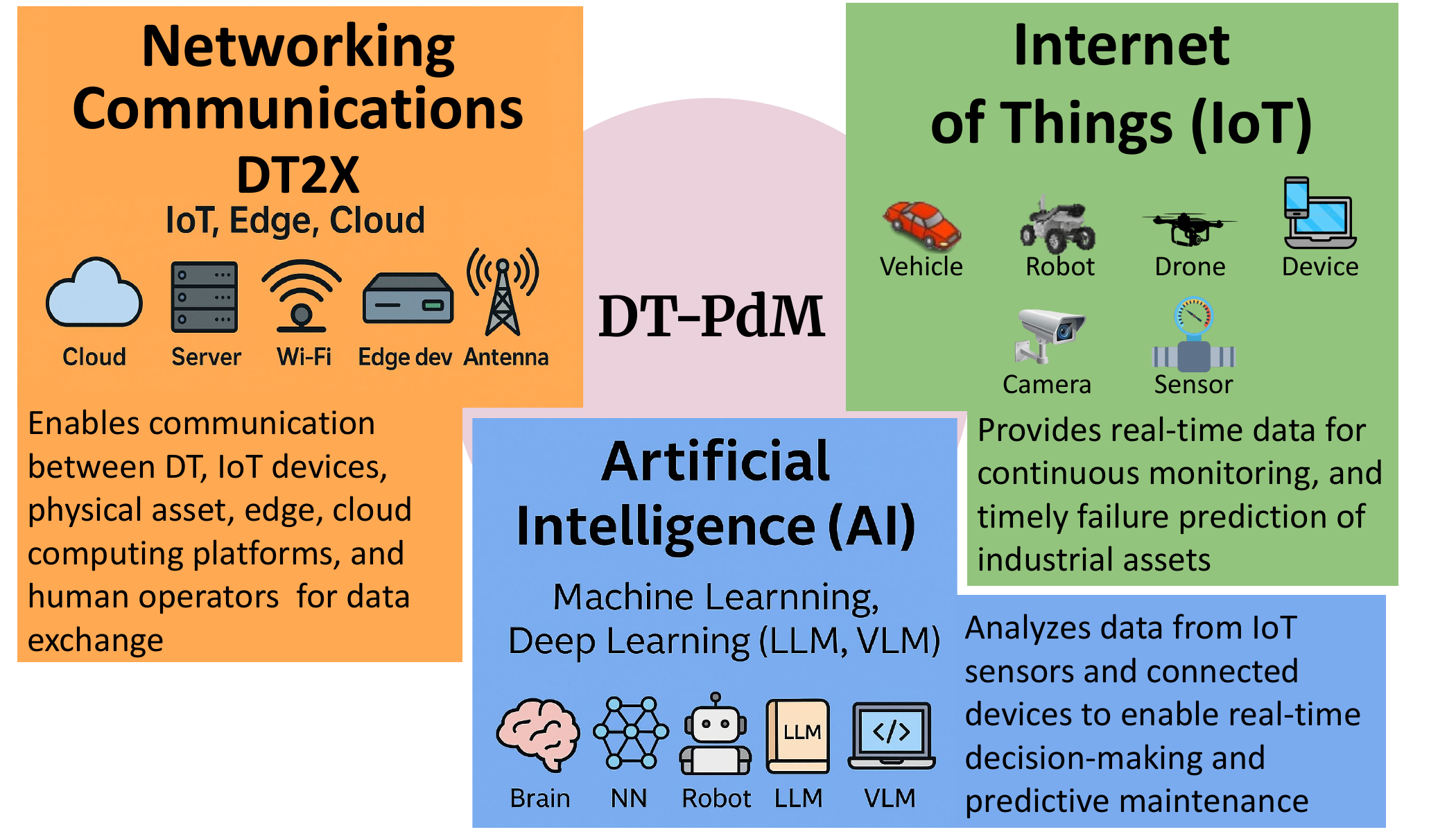}
    \caption{Emerging technologies enabling Digital Twin-based Predictive Maintenance}
    \label{fig:introductiondt}
\end{figure}

 As shown in Figure \ref{fig:introductiondt}a  DT-PdM system is underpinned by the emerging technologies of 1) Internet of Things (IoT) sensors that continuously capture real-time operational data, 2) AI-powered DT for decision-making, and 3) distributed communication based on integrated IoT, edge, and cloud technologies \cite{ismail2022artificial}.
Based on a report by Staista \cite{vailshery2024industrial}, the global market for IIoT was sized at over 544 billion U.S. dollars in 2022, and the market is expected to grow in size in the coming years, reaching some 3.3 trillion U.S. dollars by 2030. Smart factors enabled by IIoT and digitalization are emerging as a replacement for traditional factories \cite{statistaTopicSmart}. This highlights the rapid adoption of connected IoT devices and DT technologies in industrial engineering processes, such as PdM. However, for the DT-PdM to achieve its full potential, and end-to-end integrated IIoT-PdM, edge and cloud computing system architecture must be in place, enabling quality-of-services, real-time communication and decision-making. 

This paper presents a comprehensive systematic review of the state-of-the-art in DT-enabled predictive maintenance in industrial engineering. The key contributions are as follows.
\begin{itemize}
    \item We provide a temporal evolution of the DT-PdM from its conception and simulation for the Apollo 13 mission to its AI-enabled integration, providing a retrospective analysis on the requirements of the applications using the DT-PdM.
    \item We propose a taxonomy of DT-PdM-enabled industrial engineering applications based on their industrial domains. In addition, we provide insights on the requirements of these applications that the underlying DT technology should consider.
    \item We classify DT-PdM algorithms based on the underlying models used for maintenance prediction. 
    \item We devise the architectural elements for DT-PdM, enabling an end-to-end \textit{DT2X} communication, and an AI-driven decision-making system, highlighting the challenges for the realization of the system in terms of QoS, real-time, scalability, security and energy optimization.
    \item We implement a prototype demonstrator of AI-DT-PdM applications deployed in Edge and Cloud computing environments.
    \item We propose future research directions toward the realization of an efficient, real-time, scalable, trustworthy, and green DT-PdM digital ecosystem.
\end{itemize}
\begin{figure*}[!ht]
    \centering
    \includegraphics[width=0.7\textwidth]{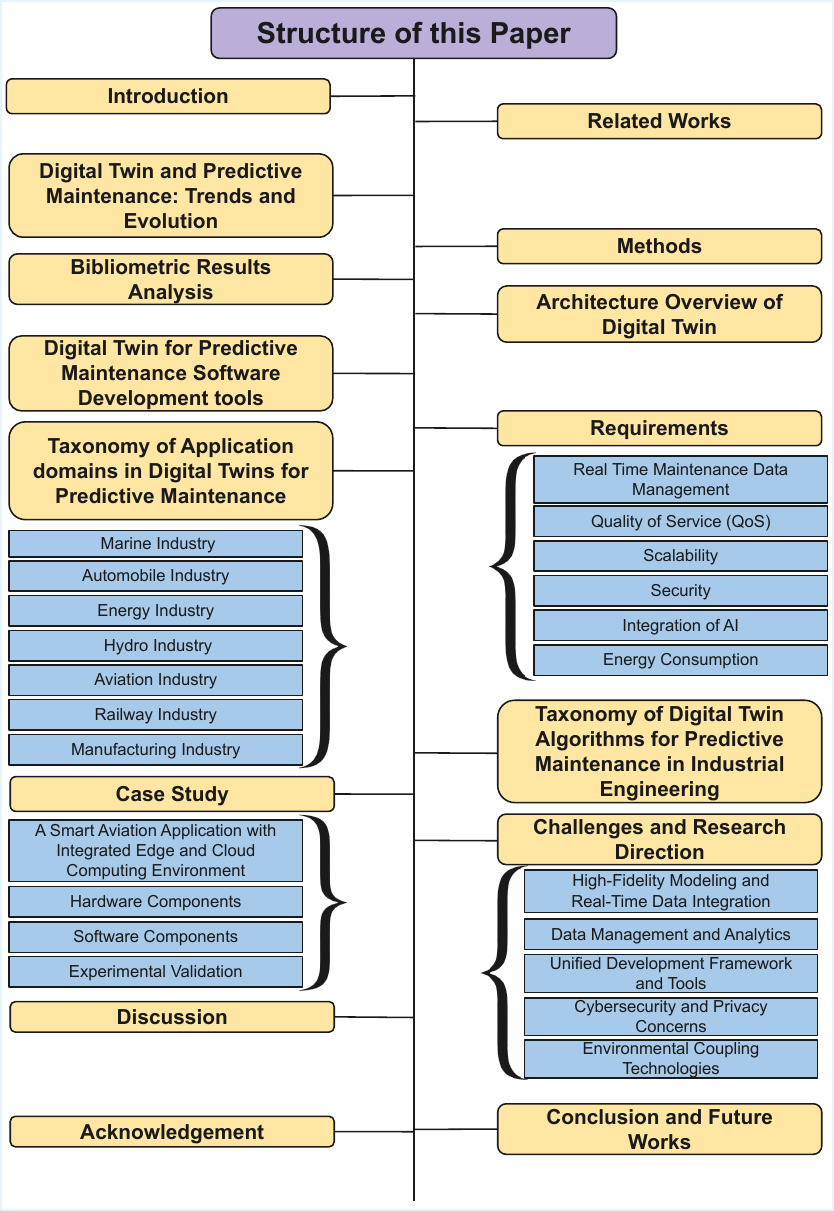}
    \caption{Organization of the review.}
    \label{fig:organization}
\end{figure*}
Figure \ref{fig:organization} shows the organization of the rest of the paper:

Section \textbf{2} provides an overview of related surveys, while Section \textbf{3} presents a retrospective analysis of the temporal evolution of the characteristics underlying the development of DT-PdM. Section \textbf{4} describes our systematic review methodology, followed by Section \textbf{5} that presents the key bibliometric results. Section \textbf{6} presents our vision of a layered architecture overview of the DT-PdM.  Section \textbf{7} highlights the existing software tools used in the literature to build DT-PdM. DT-PdM requirements are discussed in Section \textbf{8}. Sections \textbf{9} and \textbf{10} orchestrate the taxonomies of industrial engineering applications using DT-PdM, and the corresponding algorithms, respectively. Section \textbf{11} presents an application demonstrator of AI-DT-PdM, while Section \textbf{12} addresses current challenges and future directions. Discussion is provided in Section \textbf{13}, and Section \textbf{14} concludes the review.

\section{Related Works}
Several surveys in the literature \cite{18wang2023review, 19zhong2023overview, 20van2022predictive, 21falekas2021digital, ma2024state, 37chen2023advance, mayr2024digital, cimino2024enhancing} emphasize the evolution of DT concepts and their integration with PdM. However, those works do not lay out the architectural elements of DT-PdM, which are necessary for the realization of a Quality-of-Services (QoS), real-time, AI-driven, and secure DT-PdM in a distributed IIoT, edge, and cloud computing integrated system. In addition, most of these studies do not provide a systematic methodology for collecting, including, excluding, and analyzing the relevant papers under study. Table~\ref{tab:table1} summarizes the comparisons across these works, in terms of the requirements considered (e.g., real-time performance, QoS, scalability, security, AI, and energy), search strategies, and the targeted application domains. Unlike earlier reviews, our review paper not only adheres to a systematic methodology but also extends DT-PdM applications beyond traditional settings, enriching it with AI and distributed architectural elements necessary for its efficient realization. We integrate these findings by addressing a
broader spectrum of industrial domains—including marine, automobile, energy, hydro, aviation, railway, and manufacturing industries—and establishes a unified framework for implementing DT-based PdM solutions. Furthermore, our work further integrates its findings by addressing a broader spectrum of industrial domains—including marine, automobile, energy, hydro, aviation, railway, and manufacturing industries—and establishes a unified framework for implementing DT-based PdM solutions. We categorize the related reviews in the literature based on the application domains they target: aviation and aerospace. manufacturing and process equipment, and other specialized domains such as electrical machine \cite{21falekas2021digital}, sheet metal binding \cite{mayr2024digital}, internal supply chain management \cite{cimino2024enhancing}.

Regarding, Aviation and Aerospace, Wang et al. \cite{18wang2023review} provide an in-depth review of DT development specifically for aircraft maintenance, repair, and overhaul (MRO) activities, exploring both data-driven digital twin (DDDT) and model-based digital twin (MBDT) for enhanced aircraft system management. Ma et al. \cite{ma2024state} propose a systematic approach aimed at standardizing PdM automation in aviation by explicitly defining informational and functional requirements. Furthermore, Chen et al. \cite{37chen2023advance} showcase a multi-plant production planning platform that leverages simulation-based DT capabilities in aerospace contexts, emphasizing the importance of standardized data frameworks and cloud-edge collaboration.

Regarding Manufacturing and Process equipment, Zhong et al. \cite{19zhong2023overview} critique the limitations of traditional maintenance approaches, advocating for DT-based PdM that incorporates material properties, operating conditions, and degradation laws to enable improved fault prediction and more accurate estimation of remaining useful life. Complementing this perspective, Van Dinter et al. \cite{20van2022predictive} perform a systematic review that categorizes the existing literature along various dimensions such as DT platform implementation, communication protocols, and complexity challenges across manufacturing, energy, aerospace, and marine PdM applications.

In Specialized domains of electrical machines, Falekas and Karlis \cite{21falekas2021digital} focuses on predictive maintenance for electrical machines, highlighting the predominant use of AI techniques and categorizing digital replicas into Digital Model, Digital Shadow, and Digital Twin. Mayr et al. \cite{mayr2024digital} addresses the challenges in sheet metal bending applications by proposing a hybrid approach that integrates both physics-based and data-driven methods. This method uniquely combines DT predictions with production data analysis to derive process parameters that are otherwise difficult to measure directly. Cimino et al. \cite{cimino2024enhancing} introduces a simulation-based DT platform to enhance internal supply chain management in the oil and gas sector. Their modular architecture supports dynamic what-if analyses, resulting in notable improvements in system flow time and tardiness reduction.

\begin{table*}[!htb]
\centering
\caption{Comparison of existing surveys on Digital Twin for Predictive Maintenance}
\label{tab:table1}
\renewcommand{\arraystretch}{1.2}
\setlength{\tabcolsep}{4pt}
\begin{tabular}{p{0.8cm}|p{0.4cm}|p{0.2cm}p{0.2cm}p{0.2cm}p{0.2cm}p{0.2cm}p{0.2cm}|p{3cm}|p{0.8cm}|p{1.2cm}|p{1.5cm}|p{3.2cm}}
\toprule
\textbf{Work} &  & \multicolumn{6}{c|}{\textbf{Requirements Considered}} & \textbf{Search Databases} & \textbf{Search String} & \textbf{Inclusion \& Exclusion Criteria} & \textbf{Challenges Considered} & \textbf{Application Domain} \\
 & \rotatebox{90}{\textbf{Architectural Elements}} & \rotatebox{90}{\textbf{Real-Time}} & \rotatebox{90}{\textbf{QoS}} & \rotatebox{90}{\textbf{Scalability}} & \rotatebox{90}{\textbf{Security}} & \rotatebox{90}{\textbf{AI}} & \rotatebox{90}{\textbf{Energy}}&  &  &  & &  \\ 
\midrule
\cite{18wang2023review} & $\times$ & \checkmark & \checkmark & $\times$ & $\times$ & \checkmark & $\times$ & NR & NR & NR & $\times$ & Aircraft Vehicular Predictive Maintenance Systems \\ 
\midrule
\cite{19zhong2023overview} & $\times$ & $\times$ & \checkmark & \checkmark & $\times$ & \checkmark & $\times$ & NR & NR & NR & $\times$ & Process Equipment Manufacturing, Automobile Manufacturing, Cyber-Physical Systems \\ 
\midrule
\cite{20van2022predictive} & $\times$ & $\times$ & \checkmark & \checkmark & $\times$ & \checkmark & \checkmark & Springer, Scopus, IEEE Xplore, ScienceDirect, ACM, Wiley, Taylor \& Francis & \checkmark & NR & $\times$ & Manufacturing, Energy, Aerospace, Marine PdMs \\ 
\midrule
\cite{21falekas2021digital} & $\times$ & $\times$ & \checkmark & $\times$ & $\times$ & \checkmark & $\times$ & NR & NR & NR & \checkmark & Electrical machine control and Maintenance \\ 
\midrule
\cite{ma2024state} & $\times$ & \checkmark & \checkmark & \checkmark & $\times$ & $\times$ & $\times$ & IEEE Xplore, Science-Direct, Springer Link, ACM, Taylor \& Francis & \checkmark & \checkmark & $\times$ & Aviation, Manufacturing, HVAC Systems, Building Management\\
\midrule
\cite{37chen2023advance} & $\times$ & \checkmark & \checkmark & \checkmark & $\times$ & \checkmark & \checkmark & IEEE Xplore, Science-Direct, Springer Link, Scopus, Google Scholar, Taylor \& Francis & \checkmark & \checkmark & $\times$ & Manufacturing Systems (CNC, Bearings, Gearboxes), Aerospace Industry \\
\midrule
\cite{mayr2024digital} & $\times$ & $\times$ & \checkmark & \checkmark & \checkmark & \checkmark & $\times$ & Springer, Scopus, IEEE Xplore, ScienceDirect, ACM, Wiley, Taylor \& Francis & \checkmark & NR & \checkmark & Sheet Metal Bending (Salvagnini panel benders) \\ 
\midrule
\cite{cimino2024enhancing} & $\times$ & $\times$ & \checkmark & \checkmark & $\times$ & \checkmark & $\times$ & Springer, Scopus, IEEE Xplore, ScienceDirect, ACM, Wiley, Taylor \& Francis & \checkmark & NR & $\times$ & Oil \& Gas Manufacturing Sector (Internal Supply Chain Management) \\ 
\midrule
\textbf{Ours} & \checkmark & \checkmark & \checkmark & \checkmark & \checkmark & \checkmark & \checkmark & Springer, Scopus, IEEE Xplore, ScienceDirect, ACM, Wiley, Taylor \& Francis & \checkmark & \checkmark & \checkmark & Marine Industry, Automobile Industry, Energy Industry, Hydro Industry, Aviation Industry, Railway Industry, Manufacturing Industry \\
\bottomrule
\end{tabular}
\\[1ex]
\footnotesize{\textbf{Note:} NR - Not Reported, $\times$ - Not considered, \checkmark\ - Considered.}
\end{table*}
\section{Digital Twin and Predictive Maintenance: Trends and Evolution}
\begin{figure}[htbp]
    \centering
    \includegraphics[width=0.7\columnwidth]{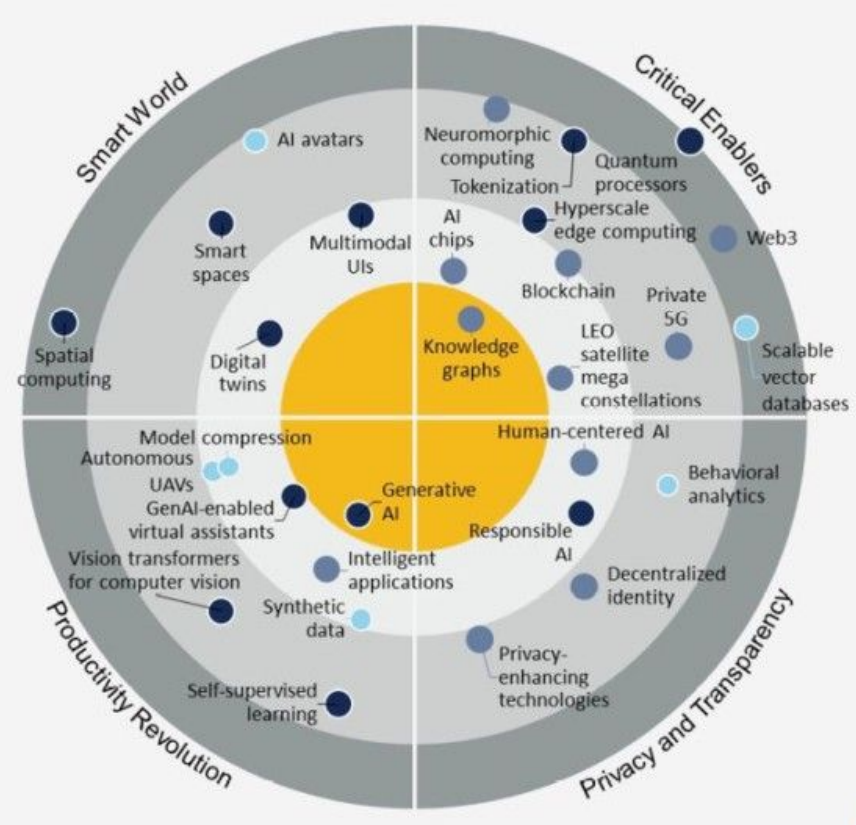}
    \caption{Emerging technologies trends (Source: Gartner Report 2024)}
    \label{fig:garttrends}
\end{figure}
\begin{figure*}[htbp]
    \centering
    \includegraphics[width=0.7\textwidth]{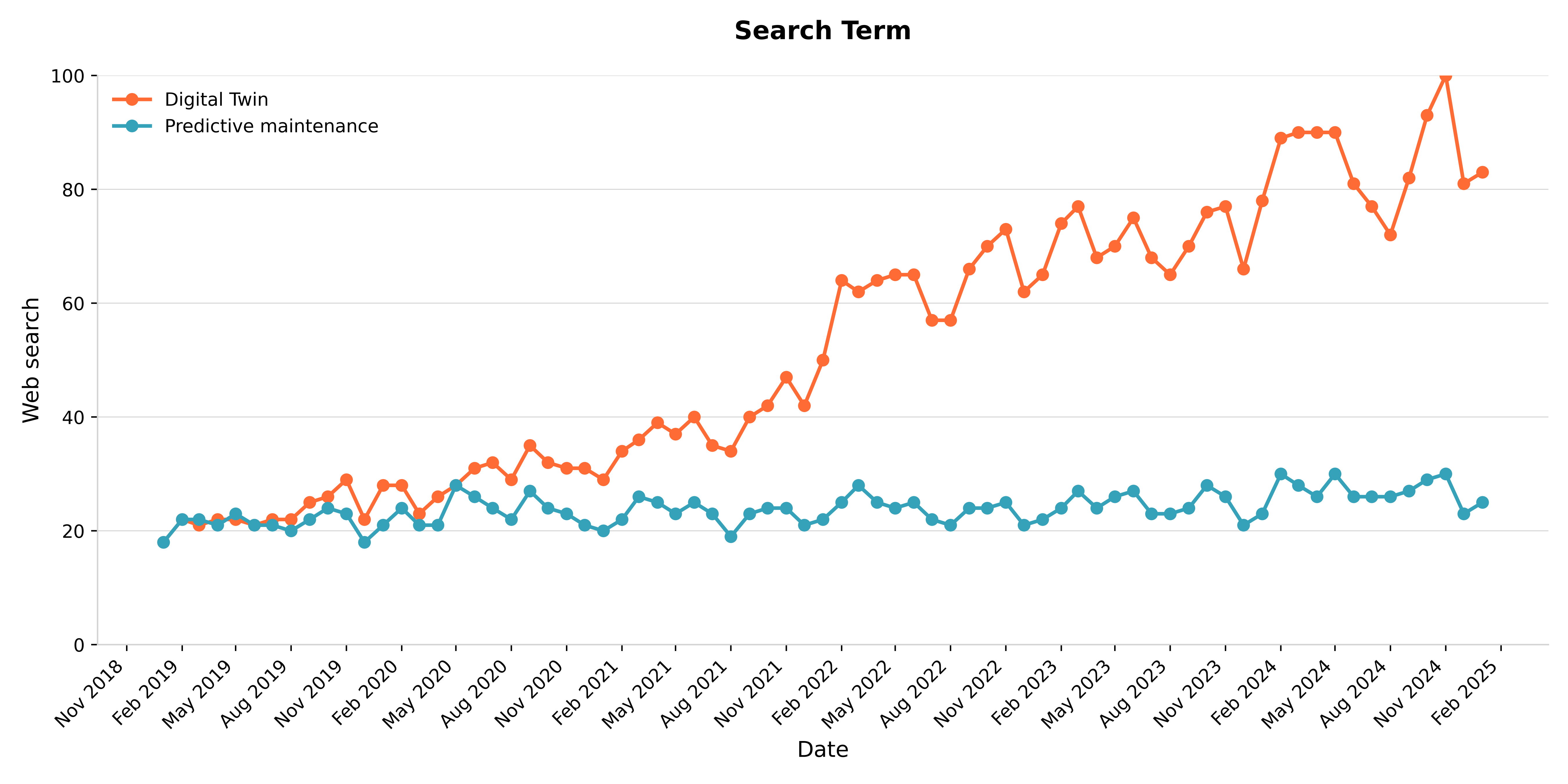}
    \caption{Google web search trends}
    \label{fig:googletrends}
\end{figure*}

During the past decade, there has been an increasing interest in DT. This domain is identified as one of the top technological trends in 2024, as shown by Gartner Radar (Figure \ref{fig:garttrends}). The radar in \ref{fig:garttrends} shows the stages over time of each emerging technology from early adoption to majority adoption by applications \cite{gartner2024}. The range in the radar measures the number of years that it will take the technology to cross from emergence to maturity. The mass represents how substantial the impact of the technologies will be on existing products and markets. The DT is projected to take 1 to 3 years for market adoption. As shown in Figure \ref{fig:googletrends}, during the last 5 years the search popularity of the DT has been increasing,
\begin{figure*}[!ht]
    \centering
    \includegraphics[width=0.8\textwidth]{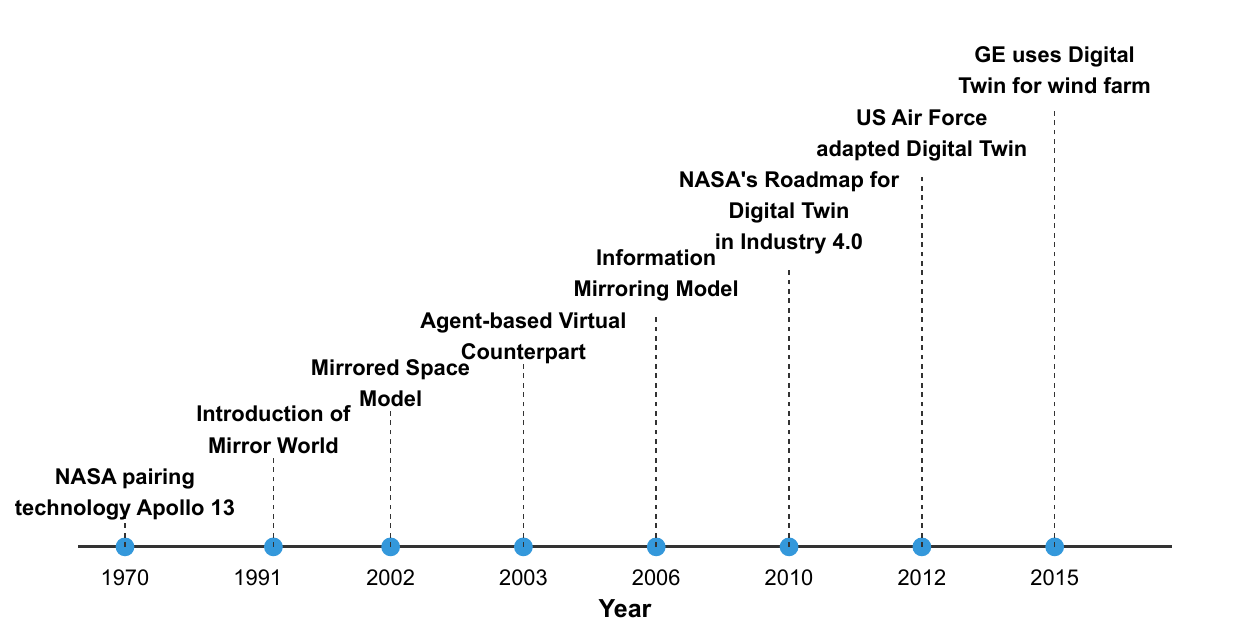}
    \caption{Evolution of Digital Twin technology}
    \label{fig:evolution}
\end{figure*}

DT technology has evolved over the years.  Figure \ref{fig:evolution} presents the timeline of the technological evolution, highlighting key definitional changes over time. The concept of a DT dates back to 1970 when NASA used a virtual replica of the Apollo 13 spacecraft to monitor its real-time conditions during the mission, saving astronauts' lives \cite{santi2023apollo13}. In 1977, a software simulation is integrated into flight physical simulators, demonstrating the capability of computational models to replicate physical systems and facilitate preparation for aerospace scenarios \cite{dimitris2022digitaltwinindustry}. However, there had been no connection between the digital and physical worlds. In 1991, "Mirror Worlds" is conceptualized, referring to software models that emulate reality through data derived from the physical environment \cite{singh2021digital}. In 2002, the framework "Mirrored Spaces Model" is introduced, consisting of real space, virtual space, and a connecting mechanism to facilitate the transfer of data and information between the two \cite{grieves2005product}. In 2003, an agent-based architecture for Product Lifecycle Management was proposed, suggesting that each product item should have a corresponding Agent "virtual counterpart" to address the inefficiencies associated with the paper-based transfer of production information \cite{framling2003product}. In 2006, the "Information Mirroring Model" replaced the "Mirrored Spaces Model". This later modification emphasizes on the bidirectional linking mechanism and the presence of multiple digital models for a single physical model.

The first comprehensive definition of a DT is announced in the 2010 NASA roadmap \cite{nasaroadmap, rosen2015importance}. NASA’s definition characterized DT as an integrated, multi-physics, multi-scale, probabilistic simulation of a vehicle or system that utilizes the best available physical models, sensor updates, and fleet history to mirror the life of its flying twin. This comprehensive definition, including the bidirectional connectivity nature between the physical and virtual worlds, enables the efficient development and PdM of physical entities in industrial engineering, and the exploration of alternative ideas or designs, such as traffic flow in Smart City, where different DT traffic flow models can be explored, and the best would be implemented in the physical world \cite{rezaei2023digital}.

With the recent emergence of IoT, big data anlaytics, AI, edge computing,  cloud computing \cite{ismail2008formal} and Privacy-Preserving Blockchain \cite{ismail2019review, hennebelle2024secure}, DT has taken wings into Industry 4.0 \cite{sepasgozar2021metrics}. With this emergence, several industries such as US Air Force (2012) \cite{tuegel2012airframe, gockel2012challenges}, GE's digital wind farm \cite{geLaunchesNext} adopted DT technologies for aircraft design and PdM.

Currently, DT is used in many fields, but mainly in the industrial sector for tasks such as fault diagnosis, PdM, evaluation and verification, real-time monitoring, production planning, and performance prediction \cite{liu2021review}. The Google search popularity trend (Figure \ref{fig:googletrends}) for DT in PdM started to emerge since 1918 with an increase in the last year, suggesting that more research and development has to be directed into that sector.
\section{Methods}
We conduct a systematic literature review search using the Preferred Reporting Items for Systematic Reviews and Meta-Analyses (PRISMA) approach, from Springer, Scopus, Web of Science (WoS), IEEE Xplore, ScienceDirect, the Association for Computing Machinery (ACM) databases, Wiley, and Taylor and Francis. Our search criteria is conducted on 6 February 2025 with no time bound. Its main objective is to retrieve all the studies which examine the integration of DT in PdM for industrial engineering. Table \ref{tab:table1} shows the search string used for each database. The relevant studies have to meet the following inclusion criteria: 1) published in the English language, 2) research papers, 3) real implementation, 2) DT in PdM, and 3) industrial engineering domains.

\begin{figure*}[!ht]
    \centering
    \includegraphics[width=0.7\textwidth]{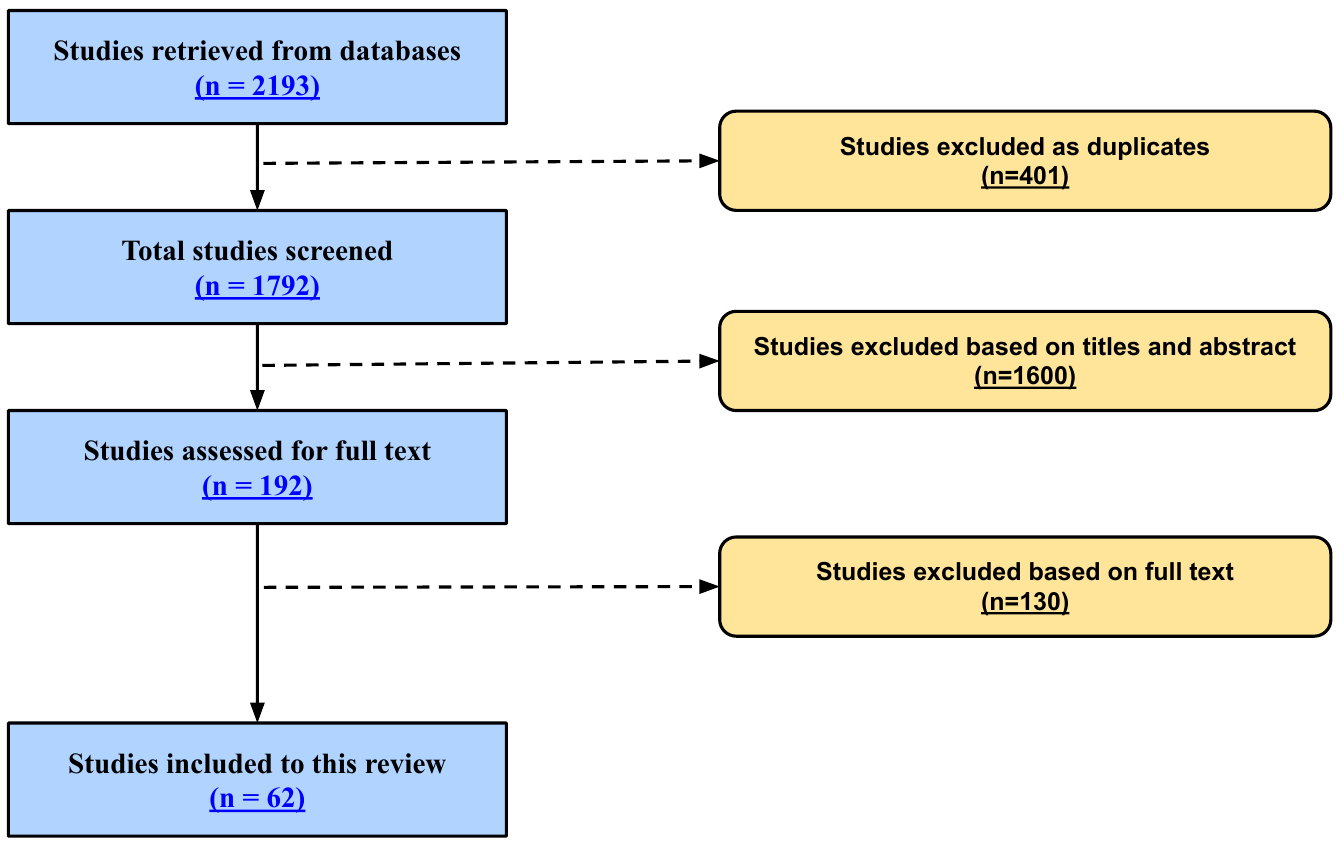}
    \caption{Flowchart of selection process to relevant studies to DT-PdM}
    \label{fig:flowchart}
\end{figure*}

\begin{table*}[!htb]
\centering
\caption{Search String used and number of retrieved articles for the databases used in this study}
\label{tab:table2}
\begin{tabular}{ll}
\toprule
\textbf{Database} &
  \textbf{Search String}\\
  \midrule
Springer &
  \begin{tabular}[c]{@{}l@{}}Where title contains - ("Digital Twin*" OR "Digital-Twin*"  OR "DigitalTwin*") AND predict* \\ AND maint* AND Industr*\end{tabular}\\
  \midrule
Scopus &
  \begin{tabular}[c]{@{}l@{}}((TITLE-ABS-KEY(Digital OR Digital-) AND TITLE-ABS-KEY(Twin*)) OR TITLE-ABS-\\ KEY(DigitalTwin*)) AND TITLE-ABS-KEY(predict*) AND TITLE-ABS-KEY(maint*)  AND \\ TITLE-ABS-KEY(Industr*)\end{tabular}\\
  \midrule
WoS &
  \begin{tabular}[c]{@{}l@{}}(TI=(((Digital OR Digital-) AND (Twin*)) OR (DigitalTwin*)) OR AB =(((Digital OR Digital-) \\ AND (Twin*)) OR (DigitalTwin*)) OR AK=(((Digital OR Digital-) AND (Twin*)) OR \\ (DigitalTwin*)))  AND (TI=predict* OR AB =predict* OR AK=predict*) AND  (TI=maint* OR \\ AB =maint* OR AK=maint*) AND (TI=Industr*  OR AB =Industr* OR AK=Industr*)\end{tabular}\\
  \midrule
IEEE Xplore &
  \begin{tabular}[c]{@{}l@{}}(((("Document Title":Digital OR "Document Title":Digital-)  AND ("Document Title":Twin OR \\ "Document Title":Twins OR "Document Title":Twinning)) OR ("Document Title":DigitalTwin)) \\ AND "Document Title":predict* AND "Document Title":maint*  AND "Document Title":Industr*) \\ OR (((("Abstract":Digital OR "Abstract":Digital-) AND ("Abstract":Twin OR "Abstract":Twins OR  \\ "Abstract":Twinning)) OR ("Document Title":DigitalTwin)) AND  "Abstract":predict* AND \\ "Abstract":maint* AND "Abstract":Industr*)  OR (((("Author Keywords":Digital OR "Author \\ Keywords":Digital-)  AND ("Author Keywords":Twin OR "Author Keywords":Twins OR  "Author \\ Keywords":Twinning)) OR ("Document Title":DigitalTwin))  AND "Author Keywords":predict* \\ AND "Author Keywords":maint*  AND "Author Keywords":Industr*)\end{tabular}\\
  \midrule
ScienceDirect &
  \begin{tabular}[c]{@{}l@{}}Title, abstract, keywords: (Digital AND Twin) AND (predict OR  prediction OR predicitve) \\ AND (maintain OR maintenance) AND  (industry OR industrial)\end{tabular}\\
  \midrule
ACM &
  \begin{tabular}[c]{@{}l@{}}((Title:(((Digital OR Digital-) AND (Twin*)) OR (DigitalTwin*))  OR Abstract:(((Digital OR \\ Digital-) AND (Twin*)) OR (DigitalTwin*)) OR Keyword:(((Digital OR Digital-) AND (Twin*)) \\ OR (Digital Twin*))) AND (Title:(predict*) OR Abstract:(predict*) OR Keyword: (predict*)) AND \\ (Title:(maint*) OR Abstract:(maint*) OR Keyword: (maint*)) AND (Title:(Industr*) OR \\ Abstract:(Industr*) OR Keyword: (Industr*))\end{tabular}\\
  \midrule
Wiley &
  \begin{tabular}[c]{@{}l@{}}(((Digital OR Digital-) AND (Twin*)) OR (DigitalTwin*)) AND  predict* AND maint* AND \\ Industr* in Title, Abstract, and Keywords\end{tabular}\\
  \midrule
\begin{tabular}[c]{@{}l@{}}Taylor and \\ Francis\end{tabular} &
  \begin{tabular}[c]{@{}l@{}}(((Digital OR Digital-) AND (Twin*)) OR (DigitalTwin*)) AND  predict* AND maint* AND \\ Industr* in Title, or Abstract, or Keywords\end{tabular}\\
  \bottomrule
\end{tabular}
\end{table*}
After the initial selection of several published papers by analyzing the title, abstract, and keywords of each article, the following inclusion criteria must be met for the articles to be considered in this review:
\begin{itemize}
    \item \textit{Inclusion criteria 1:} Focus should be on PdM using DT
    \item \textit{Inclusion criteria 2:} Should be in the domain of industrial engineering
    \item \textit{Inclusion criteria 3:} Should include implementation and performance evaluation
\end{itemize}
\section{Bibliometric Results Analysis}

Figure \ref{fig:flowchart} shows the adopted PRISMA systematic flow diagram followed for study selection. An initial investigation revealed 384 duplicate studies extracted from the databases considered. The remaining 1792 studies were screened for titles and abstracts. Consequently, 1600 studies were removed and 209 studies were further evaluated for full text. Based on the inclusion and exclusion criteria mentioned above, a total of 130 studies were excluded after full-text analysis, and the remaining 62 studies were included in this review.

\begin{figure}[!h]
    \centering
    \includegraphics[width=\columnwidth]{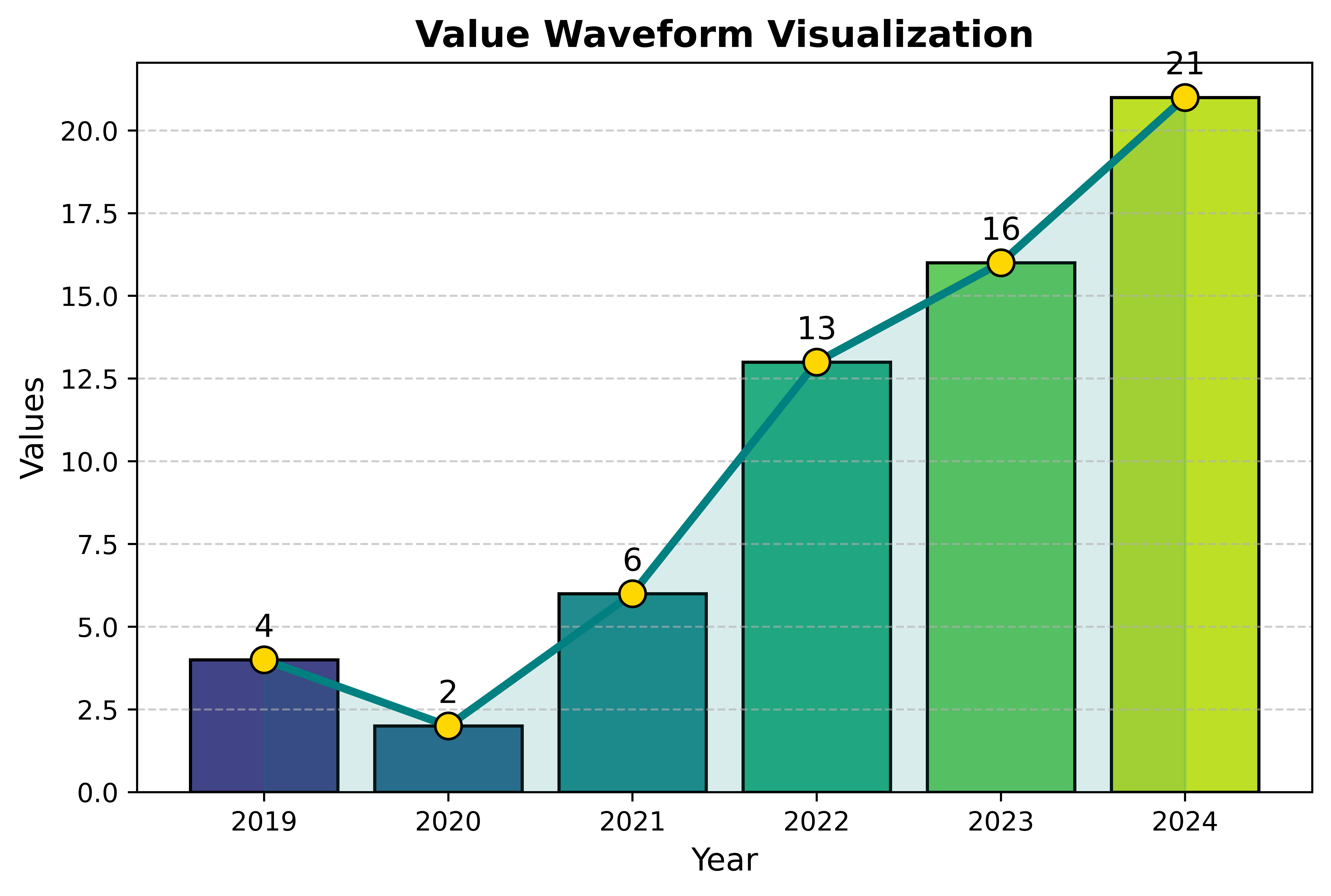}
    \caption{Number of articles included in this study per year}
    \label{fig:studyperyear}
\end{figure}

\begin{figure}[!h]
    \centering
    \includegraphics[width=\columnwidth]{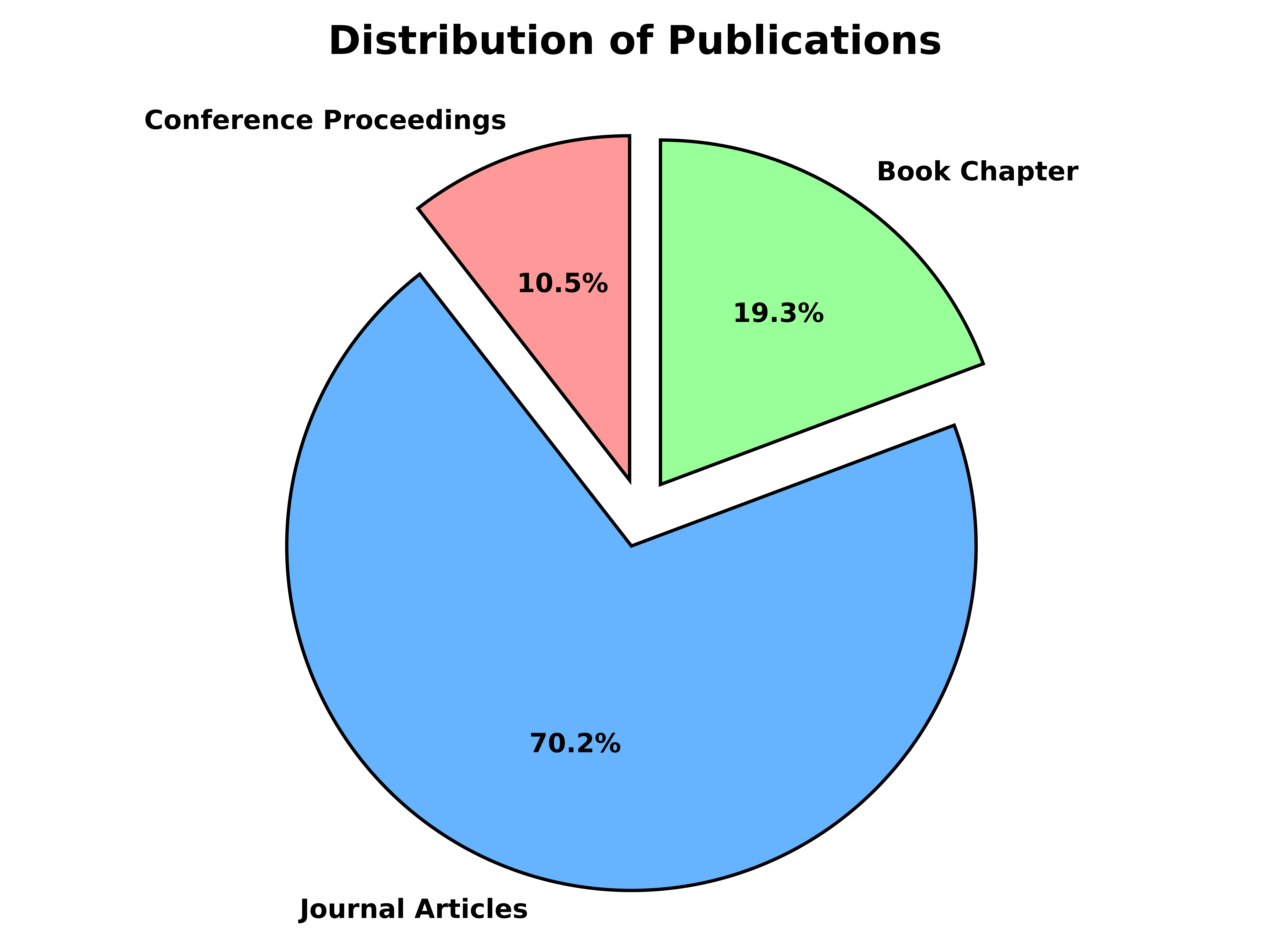}
    \caption{Types of articles in this study}
    \label{fig:types}
\end{figure}

\ref{fig:studyperyear} presents the trend of the included articles over the years. It shows that the research interest in DT for PdM has been increasing over the years. Also, most of the included articles are journal articles, followed by conference proceedings (Figure \ref{fig:types}), showing the attractiveness of the field among researchers. IN addition, almost half of the journal articles on DT-PdM implementations in industrial engineering are attributed to the manufacturing industry, demonstrating the broad and diverse application of DT technology across various machinery and processes. Only 6 of the considered articles are book chapters.

\begin{figure*}[!h]
    \centering
    \includegraphics[width=0.7\textwidth]{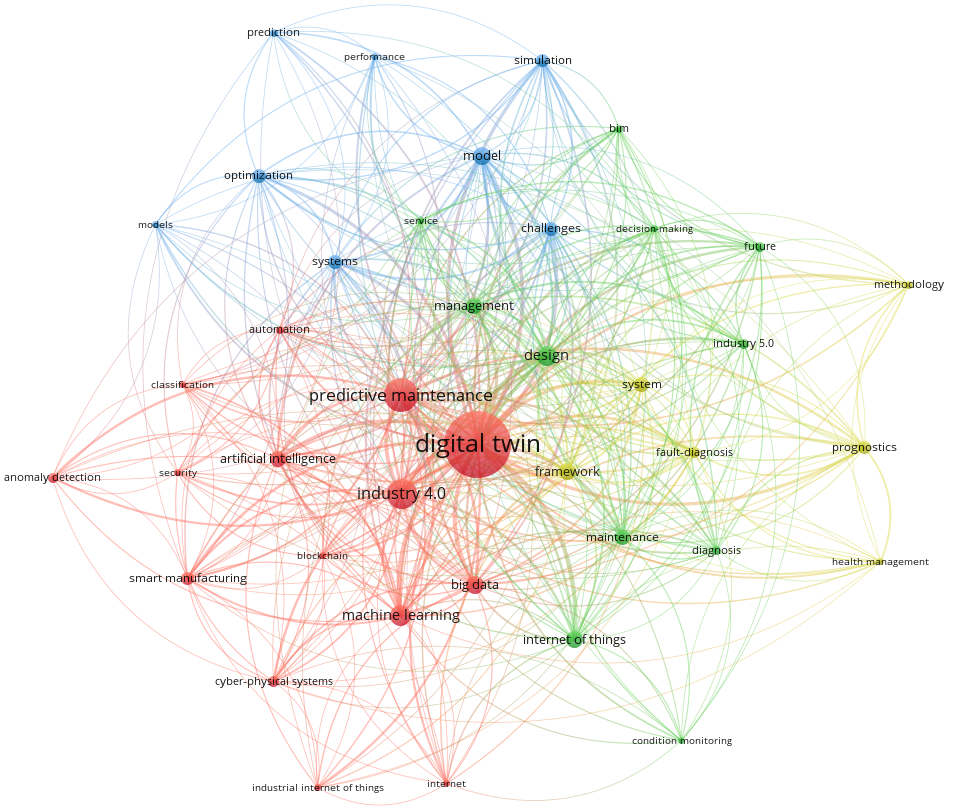}
    \caption{Relevant studies keywords network visualization}
    \label{fig:keywords}
\end{figure*}

Figure \ref{fig:keywords} shows the visualization of the network of keywords selected from the included articles. The keywords are chosen in a way that they co-occurred at least 2 times. Different keywords were used to refer to the same concept in different articles. Therefore, we modified those keywords for consistency. Table \ref{tab:table3} presents the original and replaced keywords used for visualization. The network visualization represents the keyword co-occurrences, with circle sizes indicating their weights in terms of the number of co-occurrences among the papers under study. The figure shows that the keywords are grouped into four groups. Group 1 includes AI, DT, industry 5.0, industry 4.0, IoT, and PdM, indicating a strong correlation between DT and those emerging technologies. Group 2 includes energy efficiency, industrial IoT, machine learning, and optimization, showing interest in energy management and optimization by leveraging DT and related technologies. Group 3 includes a decision support system, dynamics, a physics-based model, and the remaining useful life. Finally, group 4 includes Computer Numerical Control (CNC) \cite{luo2019digital} and fault diagnosis. These clusters highlight research interest in automation, optimization, and predictive modeling in industrial operations and maintenance.
\begin{figure}[!h]
    \centering
    \includegraphics[width=\columnwidth]{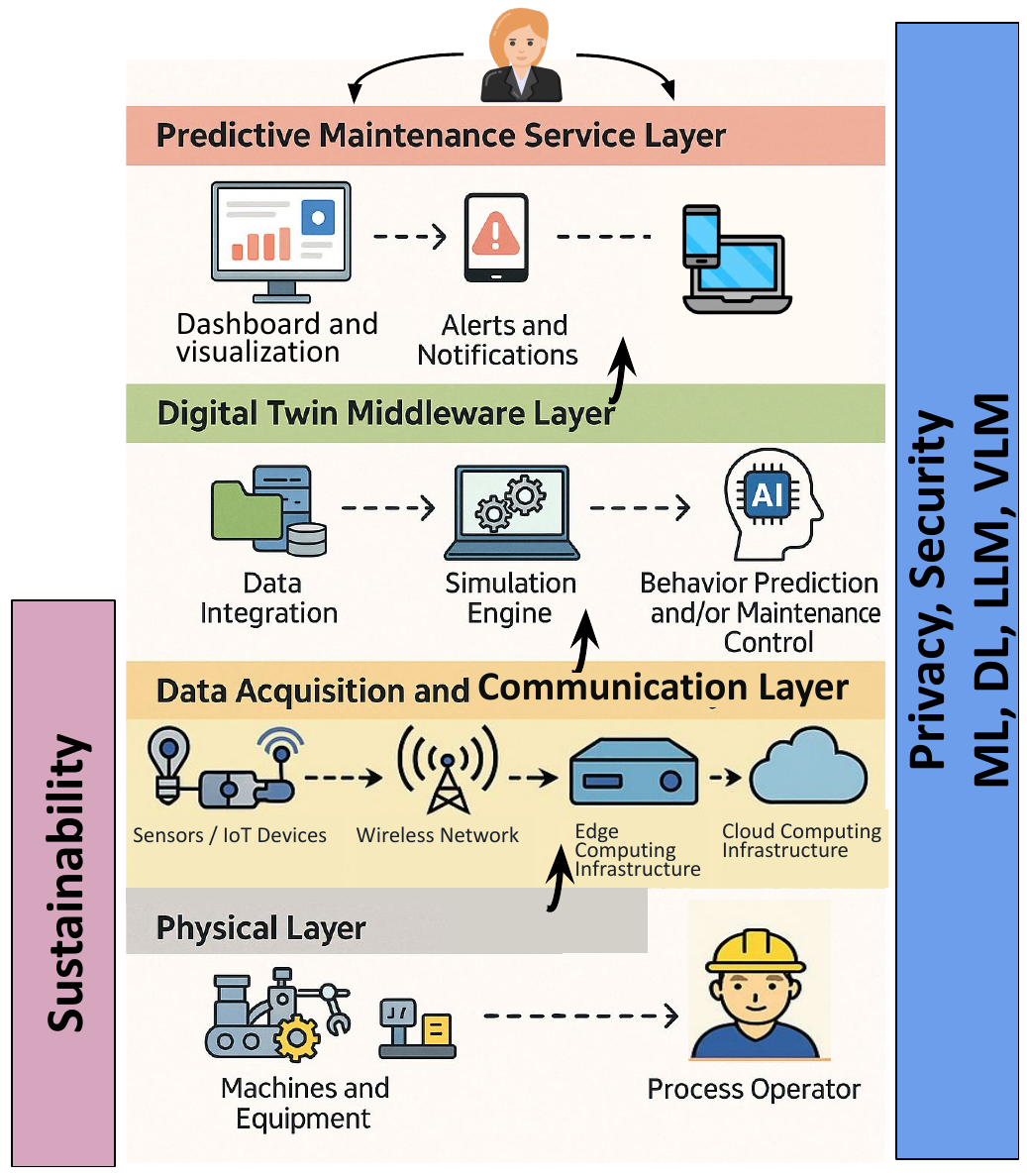}
    \caption{Layered architecture overview of the Digital Twin for Predictive Maintenance}
    \label{fig:overviewdtpdm}
\end{figure}
\begin{table*}[!h]
    \centering
    \renewcommand{\arraystretch}{1.2}
    \begin{threeparttable}
        \caption{Overlapping keywords used for visualization}
        \label{tab:table3}
        \begin{tabular}{p{0.6\textwidth} p{0.35\textwidth}}
            \toprule
            \textbf{Keywords Used in Articles} & \textbf{Keywords Used for Visualization} \\
            \midrule
            Artificial Intelligence, aritificial intelligence, ai & artificial intelligence \\
            Cyber-physical system, Cyber physical system & Cyber-physical system \\
            Digital twin, digital twin, digital twin technology, Digital twins, Digital Twins, Digital twins, Digital twins (DT), Digital Twin & digital twin \\
            Remaining useful life, remaining useful life estimation, Remaining Useful Life prediction, RUL prediction & Remaining useful life \\
            Fault diagnose, Fault Diagnosis, Fault diagnosis & Fault diagnosis \\
            Industrial Internet of Things, Industrial Internet of things (IIoT) & industrial internet of things \\
            Internet of Things, Internet of things, IoT & internet of things \\
            Machine learning, Machine Learning & machine learning \\
            Optimization, optimization & Optimization \\
            Predictive maintenance, predictive maintenance, Predictive Maintenance & predictive maintenance \\
            Energy efficiency, energy efficiency & Energy efficiency \\
            Industry 4.0, industry4, industry4.0 & industry 4.0\\
            Industry 5.0, industry5, industry5.0 & industry 5.0\\
            \bottomrule
        \end{tabular}
    \end{threeparttable}
\end{table*}

\section{Architecture Overview of Digital Twin}
\label{sec:overview}
DT technology has emerged as a critical enabler for PdM, providing real-time monitoring, simulation, and predictive analysis of physical systems. The architecture of DT for PdM revolves around integrating physical assets with digital models, analytics, and control systems to enhance asset reliability and reduce downtime.

Figure \ref{fig:overviewdtpdm} presents our vision of a four-layer architecture that underpins a DT system for PdM, ensuring efficient data flow and processing for optimal maintenance operations \cite{23tao2017digital, 24luo2020hybrid, 25wang2022digital}. The Physical Layer consists of the actual machines, equipment, and systems being monitored. The next layer is the Data Acquisition and Transmission Layer. This layer is responsible for collecting data (such as temperature, vibration, and pressure) through sensors and IIoT devices, and transmitting it to higher layers for pre-processing and analysis. Sensors and IoT devices embedded in the equipment gather operational data and relay it through a wireless network, which can include technologies, such as 5G, Wi-Fi, or LoRa, depending on the system under monitoring requirements. This data is first transmitted to edge computing, where edge devices examine the data, taking action in case of emergency, and pre-process it locally to reduce latency, before being sent to a cloud computing infrastructure. The latter plays a vital role in providing the necessary storage and computing power for large-scale data processing and analytics, and the training phase of the AI and machine learning operations, while the edge is responsible for the inference phase. Above, at the core is the DT Layer, where a virtual representation of the physical asset is modeled. This layer includes simulation engines for predictive analysis, allowing for simulations of future asset behavior, and data storage for real-time data. The predictive models are continuously updated with new data to enhance accuracy and reliability. The topmost layer, the PdM Service Layer, provides the interface through which human operators and maintenance teams interact with the DT. A dashboard offers real-time visualizations of asset performance, operational status, and predictive insights, giving maintenance teams a clear view of equipment health and performance trends. Automated alerts and notifications are generated when anomalies are detected or when predictive models forecast potential failures, ensuring that maintenance teams can take timely action.
\section{Digital Twin for Predictive Maintenance Software Development Tools}
\label{sec:tools}
Table \ref{tab:software} shows a comparison between the software tools in terms of strengths and weaknesses. Different software tools have been used in the literature to develop a DT model for the corresponding physical model. \cite{43zhang2022digital} employs the Open-Simulation Platform (OSP), while \cite{83opensimulationplatformOpenSimulation, 48singh2023building, 49guc2022smart, 72laura2019supportmachine, 75Selcuk2021synthetic, 44szpytko2021digital, 51amadou2023energy, 59unal2022data, 62zayed2023efficient, 65feng2023digital, 69siddiqui2023artificial, 70davies2022digital, 71mi2021prediction} use MATLAB Simulink. \cite{45rajesh2019digital} experiments with CREO Simulate software, \cite{84creo, 56padovano2021prescriptive} utilizes AnyLogic, \cite{52wang2022digital} use Unigraphics NX (UG) software to develop a 3D equipment model and Ansys Mechanical software to find model parameters, \cite{54sresakoolchai2023railway} experiments with Autodesk Civil 3D, \cite{57aivaliotis2019use} uses OMEdit environment. \cite{85openmodelicaOMEditOpenModelica, aivaliotis2021degradation, 63aivaliotis2019methodology, 64aivaliotis2023methodology}, utilize OpenModelica and MATLAB. \cite{50haghshenas2023predictive}, \cite{46eaty2023digital}, \cite{61panagou2022explorative}, and \cite{80suve2021predictive} adopt Unity3D, Microsoft Azure Cloud Platform, Eclipse Ditto, and Flexsim, respectively, and \cite{81meng2023prediction} use LS-DYNA to develop a DT model. On the other hand, works \cite{47deon2022digital, 53heim2020predictive, 55wang2021complex, 58luo2020hybrid, 60ren2023edge, 66hassan2024experience, 68panagou2022feature, 67banyai2022real, 74yu2023dynamicmodel, 76pecora2023monitoringmachine, 77wei2022augmented, 78xue2022cncmachine, 79Borangiu2023robothealth, 82lian2022application} did not report the software they used to create a DT. 

\begin{table*}[htbp]
\label{tab:software}
\centering
\scriptsize
\renewcommand{\arraystretch}{1.3}
\setlength{\tabcolsep}{2pt}
\caption{Comparison of simulation software tools used in DT-PdM in the literature}
\begin{tabular}{@{}p{2.2cm}p{4.4cm}p{2.7cm}p{2.7cm}p{1.5cm}@{}}
\toprule
\textbf{Tool} & \textbf{Description} & \textbf{Strength} & \textbf{Weakness} & \textbf{References} \\
\midrule
MATLAB Simulink &
A MATLAB-based graphical programming environment for modeling, simulating, and analyzing multidomain dynamical systems &
High-speed simulations &
Can become numerically unstable with a large number of variables &
\cite{simulink} \\

CREO Simulation &
Used to analyze and validate the performance of 3D virtual prototypes &
Suitable for projects that involve complex geometries &
Poor rendering and difficulty in importing sketches &
\cite{creo} \\

AnyLogic &
Modeling tool for agent-based, discrete event, and system dynamics simulation &
Flexible and easy to understand &
Slow performance &
\cite{anylogic} \\

Unigraphics NX &
Advanced CAD/CAM/CAE software for 3D simulations &
Highly configurable &
Performance slows down while simulating large parts &
\cite{nx} \\

Autodesk Civil 3D &
Used to create 3D civil designs like highways, junctions, roundabouts, railways, land developments, and storm drainage networks &
Easy to use &
Performance slows down while handling large amounts of topographic data &
\cite{civil3d} \\

OpenModelica &
Open-source environment based on the Modelica modeling language for modeling, simulating, optimizing, and analyzing complex dynamic systems &
Suitable for systems with significant non-linear components &
Complex user experience &
\cite{openmodelica} \\

OMEdit &
Graphical interface for graphical modeling in OpenModelica &
Suitable for systems with significant non-linear components and simple interface &
Not available. &
\cite{omedit} \\

Unity 3D &
Used to create immersive and interactive 3D experiences &
Cross-platform compatibility, powerful rendering, and rich asset store &
High memory consumption and not suitable for complex simulations &
\cite{unity} \\

Microsoft Azure &
Cloud service to develop and deploy software models &
Data security, high availability, and scalable &
Requires expertise &
\cite{azure} \\

Open-Simulation Platform &
Open-source software for co-simulation of maritime equipment, systems, and entire ships &
Modular and flexible &
Not available. &
\cite{osp} \\

Eclipse Ditto &
Open-source framework to build digital twin of devices connected to the internet for monitoring and controlling real-world objects &
Standardized API, scalable, flexibility in device connectivity, highly secure, stores historical data, and real-time monitoring &
Not available. &
\cite{ditto} \\

Flexsim &
Used for 3D simulation, model building, model analysis, and optimization &
Good customization and capable library of class objects &
Limited documentation, poor user interface, and connections between objects can be difficult &
\cite{flexsim} \\

LS-DYNA &
Used for 2D and 3D simulation of different sensors and building elements. It is used in oil and gas industry, military and defense, aerospace, and automotive industry &
It can accurately handle non-linear problems and non-linear materials &
High computational complexity and takes a long time to model &
\cite{lsdyna} \\
\bottomrule
\end{tabular}
\end{table*}
\section{Requirements}
\label{sec:requirements}
To enable the large-scale adoption of DTs for PdM in industrial engineering, the following requirements have to be fulfilled.
for PdM solutions
\subsection{Real Time Maintenance Data Management}
One of the key goals of DT in PdM is to ensure QoS of IIoT applications, and real-time monitoring and decision-making. Real-time capability refers to a ability of DT to mirror the physical system state and behavior with minimal latency, a critical requirement for time-sensitive applications, such as PdM \cite{26uhlemann2017digital}. For instance, in the context of floating offshore wind turbines, a diagnostic DT can monitor operational data in real-time, detect anomalies, and diagnose failures, thereby enabling condition-based and PdM. In these contexts, rapid detection of potential failures is essential to prevent costly downtime or accidents. Achieving real-time performance requires the adoption of advanced technologies and architectures to support effective data management within the DT system. This includes leveraging robust IIoT infrastructure, 5G networks, cloud-edge computing, and sophisticated data transmission protocols \cite{ismail2022artificial}. 5G, in particular, offers ultra-low latency, allowing near-instantaneous data transmission from sensors within physical systems to their digital counterparts. Cloud-edge computing further reduces latency by processing data closer to the source, eliminating the need to transfer it to distant cloud servers \cite{27li2020scheduling}. Jianping Huang \cite{28huang2021application} designed an intelligent power system that leverages situational awareness and virtual reasoning to overcome challenges related to uncertainty and errors in the power industry. This system enables real-time interaction between the virtual and physical environments to enhance problem-solving capabilities. Without real-time synchronization between the physical and DT, PdM could miss anomalies or fail to make timely recommendations, ultimately increasing the risk of equipment failure.
\subsection{Quality of Service (QoS)}
Ensuring high Quality of Service (QoS) is crucial for DT in PdM to deliver accurate and reliable data. Key QoS parameters such as availability, reliability, and latency must be rigorously maintained. A well-functioning DT relies on precise data; any inaccuracies can lead to faulty predictions, compromising the overall effectiveness of the system. In addition to accuracy, high availability is a fundamental requirement, particularly for critical systems like industrial machinery, power plants, and medical devices. The DT must remain operational at all times, as even brief downtime could result in missed fault detection, leading to costly failures. Redundancy strategies, including fail over systems and backup servers, are essential to maintaining both availability and reliability \cite{29li2022enhanced}. Low latency is equally important for PdM, ensuring that data exchanged between the physical system and its twin is processed swiftly and acted upon in a timely manner. Additionally, machine learning techniques can continuously refine predictive models, further enhancing their accuracy over time. By prioritizing QoS, DT can provide reliable and trustworthy maintenance recommendations, enabling organizations to make informed decisions and optimize asset performance.
\subsection{Scalability}
The scalability of DT in PdM is essential as systems grow and evolve. It refers to the ability of the DT architecture to handle increasing data volumes, a larger number of physical assets, and expanding operational demands without compromising performance. As industries expand, the volume of data generated by sensors and devices rises significantly \cite{ismail2022artificial}. A scalable DT must be able to manage this growth seamlessly, ensuring performance is maintained \cite{30khan2022digital}. Cloud computing contributes to scalability by providing on-demand, dynamic storage and computational resources. These technologies distribute the processing load across multiple nodes, enabling the system to efficiently manage the influx of data from large-scale industrial environments \cite{ismail2015implementation}. Furthermore, containerization and microservices-based architectures enhance scalability by allowing individual DT components to be deployed independently and scaled according to the specific computational needs of different maintenance tasks or devices. In this way, a scalable DT can effectively accommodate increasing complexity as industries grow, without the need for a complete system overhaul.
\subsection{Security}
In sectors such as aerospace, healthcare, and manufacturing, security is a critical requirement for DT in PdM, where ensuring data integrity and confidentiality is essential. As analyzed by Alcaraz et al. \cite{31alcaraz2022digital}, the security of DT systems involves both the physical devices and the digital infrastructure, necessitating robust, multi-layered security measures. Key vulnerabilities include unauthorized access to the DT or its data, cyber attacks on communication channels, and data tampering all of which could result in inaccurate predictions or even malicious manipulation of physical systems. However, based on the literature reviewed in this article, studies focused on DT security are relatively scarce.
    
To mitigate these risks, encryption methods and secure communication protocols are vital. Access control mechanisms, including role-based access control (RBAC) and attribute-based access control (ABAC), are critical in restricting who can access and modify data within the DT system. Moreover, blockchain technology offers a promising solution for maintaining data integrity by offering a tamper-proof ledger that logs every transaction or interaction with the DT. Kanak et al. \cite{32kanak2019visionary} proposed a blockchain-based distributed DT model by integrating blockchain with DT principles such as Xby-design and XaaS. This concept employs decentralization to improve the security, integrity, and accountability of the DT system. Authors in \cite{33thakur2023effective} developed an authentication scheme for cloud computing that uses blockchain to overcome security issues in DT data management. Their proposed architecture features an advanced, tri-factor authentication scheme that protects DTs from a variety of harmful threats. Putz et al. \cite{34putz2021ethertwin} have presented EtherTwin, a blockchain-based solution for DT information management that is specifically tailored to suit the security requirements of DT systems. In addition to these protections, regular security audits and penetration testing are required to discover and repair any vulnerabilities, assuring the DT system continuous safety and reliability throughout PdM.
\subsection{Integration of Artificial Intelligence} The integration of AI into DT is crucial for enhancing PdM capabilities. AI allows DT to process vast amounts of data and derive valuable insights for predicting the behavior and future condition of assets. This integration provides a more accurate, automated decision-making process that significantly improves the effectiveness of maintenance strategies. One of the primary AI-driven approaches in DT is the use of ML algorithms. These algorithms enable the DT to analyze data and identify patterns that signal impending failures. As more data is collected, ML models continuously update and refine their predictions, improving accuracy over time. This dynamic learning capability is essential in PdM, where conditions are constantly changing, and systems must adapt quickly to new information \cite{35qi2018digital}.

Another essential application of AI in DT is anomaly detection. AI algorithms, especially those based on deep learning techniques, can detect subtle deviations from normal operational patterns that may not be visible through traditional monitoring methods. This capability allows the DT to identify early signs of equipment malfunction, providing a valuable opportunity for maintenance teams to intervene before a breakdown occurs \cite{36lee2015cyber}. Moreover, AI enhances decision support in DT by offering predictive analytics, optimization, and prescriptive recommendations. These capabilities empower maintenance teams to make informed decisions about repair and replacement schedules, thereby optimizing the performance of the asset and extending its life-cycle. AI-driven insights allow for a shift from reactive to proactive maintenance strategies, ultimately reducing operational costs and improving system reliability \cite{37chen2023advance}. Integrating AI into DT for PdM requires powerful computational resources and advanced algorithms capable of processing complex datasets in real-time. This combination of AI with DT not only improves predictive accuracy but also helps automate many of the processes that would otherwise require manual oversight, making maintenance more efficient and responsive to real-time changes in asset conditions \cite{38tao2019digitaltwin}.
\subsection{Energy Consumption} Energy consumption is an increasingly important consideration in the development and deployment of DT, especially in the context of PdM, where the environmental impact and operational efficiency are key factors. DT, which continuously monitor, simulate, and predict the behavior of physical assets, can be resource-intensive, consuming significant amounts of energy during data collection, processing, and analysis. One approach to minimizing energy consumption in DT is the use of energy-efficient sensors. These sensors are responsible for collecting real-time data from physical assets, and their design can greatly influence the overall energy footprint of the DT system. Low-power sensors, which utilize advanced technologies to optimize energy usage, can significantly reduce the energy demands of the system, especially when deployed on a large scale \cite{39li2021energy}.

The optimization of computational resources is another critical factor in reducing energy consumption. DTs often require substantial computing power to process large datasets and run complex simulations. To address this, energy-efficient algorithms can be implemented to streamline data processing and simulation tasks. These algorithms are designed to reduce computational overhead, thereby minimizing energy consumption without compromising the accuracy or speed of predictions \cite{40qadir2024digital}.
    
In real-time perception environment, it is also possible to optimize both energy consumption and carbon emissions during the maintenance process. Lin et al. \cite{41lin2019concept} presents a DT system architecture that uses data analysis to identify patterns, enabling PdM and significantly reducing the overall system energy consumption.
    
In addition, the use of edge computing, which processes data closer to the source, can help to reduce the energy consumption associated with data transmission and centralized computing \cite{42bharany2022energy}. Moreover, when DT relies on cloud infrastructure for their operations, the selection of energy-efficient cloud computing solutions becomes vital. Green cloud computing, which prioritizes sustainability by optimizing energy usage in data centers, can significantly reduce the energy consumption of DT systems \cite{42bharany2022energy}. By focusing on energy-efficient design across sensors, algorithms, and computing infrastructure, industries can leverage DTs for PdM in a way that not only enhances operational efficiency but also supports broader sustainability goals. The reduction of energy consumption in DT is not just a technical challenge but also an important step towards achieving greener, more sustainable industrial practices.
\section{Taxonomy of Application Domains in Digital Twin for Predictive Maintenance}
Figure \ref{fig:taxonomyindustrydtpdm}, shows the taxonomy of industrial engineering applications which are enabled by DT technology for PdM. These applications are categorized into industrial domains as follows.

\subsection{Marine Industry}
In the context of the marine industry, Zhang et al. \cite{43zhang2022digital} discussed the development of DT for a research vessel Gunnerus in Norway to provide an integrated view of the ship’s physical assets and behavior at different stages of operation. The authors demonstrated the applicability of the DT-ship system for harbor docking, PdM, and remote crane operations. A DT is used for PdM to detect deviations in engine operating conditions and predict the engine’s remaining useful life. Szpytko and Duarte \cite{44szpytko2021digital} proposed a maintenance decision-making model for cranes operating in a container terminal to improve cranes’ operational efficiency by optimizing the maintenance scheduling process. The authors modeled a container terminal containing 10 gantry cranes, 2 power transformers, 2 emergency generators, and 2 transmission lines. They provided mathematical modeling for vessel demand, gantry cranes, generators, transformers, transmission lines, and risk indicators in terms of crane inefficiency.
\subsection{Automobile Industry}
Regarding the automobile industry, P.K. et al. \cite{45rajesh2019digital} developed a DT model for an automotive braking system consisting of a single wheel and a brake booster of a commercial passenger vehicle. The brake pressure at different vehicle speeds was measured using the ThingWorx IoT platform to simulate the prediction of brake wear for maintenance of brake pads. The authors used RStudio to perform a single tail ‘t’ test for further analysis of the simulated data. Eaty and Bagade \cite{46eaty2023digital} proposed a system to evaluate a vehicle battery’s state of charge (remaining battery capacity) and state of health (battery’s present health). In the proposed system, the DT model of a battery management system is hosted in a cloud. Based on the data from the physical management system, the state of health is predicted in the cloud while the state of charge is estimated on the management system.
\subsection{Energy Industry}
For PdM in the energy industry, Deon et al. \cite{47deon2022digital} proposed a DT-based system for fault classification in the power generation process of a thermoelectric complex involving two identical power plants. A genetic algorithm is used to tune DT parameters. Singh et al. \cite{48singh2023building} proposed a DT model for squirrel cage induction motors to predict remaining useful life and faults. The authors used the dSPACE MicroLabBox controller to integrate the experimental setup into the digital space. The DT model is used to generate time series data for remaining useful life prediction. In addition, the DT model is used to simulate categorical data with healthy and fault states. Guc and Chen \cite{49guc2022smart} proposed a DT-based approach for PdM. The authors evaluated the proposed approach for an RF impedance matching and mechatronic test bed. Multiple DTs are generated by manipulating fault severity levels, combining faults, and incorporating different environmental uncertainties. The simulated data is used to train fault detection machine learning models. Haghshenas et al. \cite{50haghshenas2023predictive} proposed a DT-based platform to detect potential failure of wind turbine components. In addition, the authors presented an augmented reality-based 3D visualization system to enhance user experience. The proposed system is evaluated on the Hywind Tampen Floating Wind Farm Configuration. Adamou and Alaoui \cite{51amadou2023energy} proposed a DT-based PdM method for induction motors where the motor losses are related to the motor faults using an energy-efficiency physics-based model.
\subsection{Hydro Industry}
In the hydro industry, Wang et al. \cite{52wang2022digital} developed a DT-based PdM model for turbines, generators, oil-immersed transformers, and hydraulic turbine governors in smart water conservancy projects. The proposed system uses data from DT to predict the remaining life of these components and make maintenance decisions. 
\subsection{Aviation Industry}
In the aviation industry, Heim et al. \cite{53heim2020predictive} proposed DT-based approaches to predict the remaining useful life of aircraft parts using two aircraft datasets. DT is used to enter the details of aircraft components and the probability of failure for that component will be displayed to users. 
\subsection{Railway Industry}
In the railway industry, Sresakoolchai and Kaewunruen \cite{54sresakoolchai2023railway} proposed a DT solution to improve the efficiency of railway maintenance. The authors used deep reinforcement learning to choose which maintenance activity to perform to improve overall efficiency. However, DT is used only for data storage, and no scenario simulations are performed. Wang et al. \cite{55wang2021complex} used DT for manufacturing and operation maintenance processes. The authors demonstrated the case of an Electic Multiple Units (EMU) bogie for PdM
\subsection{Manufacturing Industry}
For PdM in manufacturing industries, Padovano et al. \cite{56padovano2021prescriptive} propose a prototype of a DT-based decision support system for providing prescriptive maintenance in production, planning, and control. The authors presented a case study for a manufacturing plant located in Southern Italy producing turbo machinery components for the oil and gas industry. The machine and tool monitoring module within the proposed system receives and preprocess real-time sensor data from machines. The predictive analytics module then the data for anomaly detection and fault prediction using machine learning developed in MATLAB. The real-time data is then sent to the scenario analysis and testing module which simulates different production and maintenance processes using a DT. The recommender module then proposes optimal maintenance and production schedules based on the simulation results. Aivaliotis et al. \cite{57aivaliotis2019use} proposed a DT-based approach to calculate the remaining useful life of machine equipment by utilizing physics-based simulation models. The proposed approach begins with physics-based modeling of the machine under study. The DT of the real machine is simulated and fine-tuned using the data collected by the machine’s controller. The DT model is then used to simulate scenarios based on inputs from the physical model. The remaining useful life of the machine is then calculated based on the real-world and simulated outputs. The authors implemented the proposed approach to predict the useful life of a six-axis robotic structure used for welding in the assembly of a thermosiphonic system. Luo et al. \cite{58luo2020hybrid} proposed a PdM approach for CNC machine tools using the DT model and DT data. The proposed approach first develops a DT model that represents the real system. Next, historing sensing data after preprocessing (noise reduction, feature extraction, and feature selection) is used from the real system to predict the remaining useful life of the component. The prediction results by the data-driven approach are then used as observation values of a system to revise results by theoretical empirical deduction using a hybrid particle filtering algorithm. The authors evaluated the proposed approach to predict the life of a cutting tool. Unal et al. \cite{59unal2022data} proposed a framework for PdM that supports hybrid and cognitive DT. Hyrbid DT is based on data-driven DT and physical models, whereas a cognitive twin is an extension of a hybrid twin for enabling cognitive features for sensing complex and unpredicted behavior. The authors implemented the proposed approach for improved decision-making for the Spiral Welded Steel pipe production system.

Ren et al. \cite{60ren2023edge} proposed a three-level DT model for prognostic health management of process manufacturing systems. The unit-level DT modeled for each unit is constructed on edge devices for real-time monitoring and analysis of device status. The station-level DT deployed at fog devices is used to optimize and control manufacturing process parameters. The shop-level DT in the cloud is used for production planning. The proposed system is evaluated for a chemical fiber production plant. Panagou et al. \cite{61panagou2022explorative} used a hybrid approach for PdM of a hot rolling mill line in the steel-making production process by incorporating the data from a real system as well as simulated data from DT. The authors presented a cloud-based platform for supporting maintenance decisions. Zayed et al. \cite{62zayed2023efficient} proposed a DT-based fault diagnosis framework for industrial control systems that use DT to generate fault data and then use machine learning for fault prediction. The proposed approach uses the Flower pollination algorithm (FPA), PSO, Harris hawk optimization (HHO), Jaya algorithm (JA), Gray wolf optimizer (GWO), and Salp Swarm algorithm (SSA) for feature selection. Aivaliotis et al. \cite{63aivaliotis2019methodology} proposed a methodology for enabling DT using advanced physics-based modeling. The authors implemented the proposed methodology for an industrial robot. Later, Aivaliotis et al. \cite{aivaliotis2021degradation , 64aivaliotis2023methodology} proposed a DT-based methodology to represent the future behavior of a robot based on kinematics, structural characteristics, as well as degradation of mechanical parts. The authors use the DT model to simulate different scenarios for the estimation of remaining useful life. Feng et al. \cite{65feng2023digital} proposed a DT method to monitor and assess the degradation of gear surfaces. The proposed method is used to predict the remaining useful life of a gear. To evaluate the model the authors developed a DT model based on the spur gearbox test rig at the University of New South Wales. The optimal parameters for the DT model are obtained using a multiobjective grasshopper optimization algorithm. To generate degradation data for developing prediction models, degradation models are used to produce signals with degradation characteristics. Hassan et al. \cite{66hassan2024experience} proposed a DT-based PdM approach where the discrepancies between physical and digital units are used to detect the maintenance needs of process machinery and tools in the steel industry.

Bányai and Bányai \cite{67banyai2022real} developed a DT-based real-time policy model and optimization algorithm. The maintenance policy is based on real-time and forecasted (using DT) failure and operational data. The mathematical description of the approach uses the Markov decision process, Howard’s policy iteration technique, value determination equations, and evolutionary heuristics for policy optimization. Panagou et al. \cite{68panagou2022feature} used DT to simulate states and conditions to determine which data sensors can provide reliable maintenance-related predictions. The selected sensors are then used for maintenance detection using real data and data collected through DT simulation. The authors evaluated the approach by creating a DT of a rolling mill in the production line of the steel-making industry in Italy. Siddiqui et al. \cite{69siddiqui2023artificial} proposed a DT-based PdM approach to detect anomalies in an industry automation system to prevent failure. Real data from sensors related to smart industrial automation systems is used to validate the approach. Davies et al. \cite{70davies2022digital} proposed a DT model for three critical components of the milling machine tool, namely the spindle system, servo motor, and linear axis for remaining useful life prediction. Mi et al. \cite{71mi2021prediction} proposed a DT model for bearings in grinding rolls of a vertical mill for PdM. The authors used NSGA-II to optimize the DT model parameters.

\begin{figure*}[htbp]
    \centering
    \includegraphics[width=0.9\textwidth]{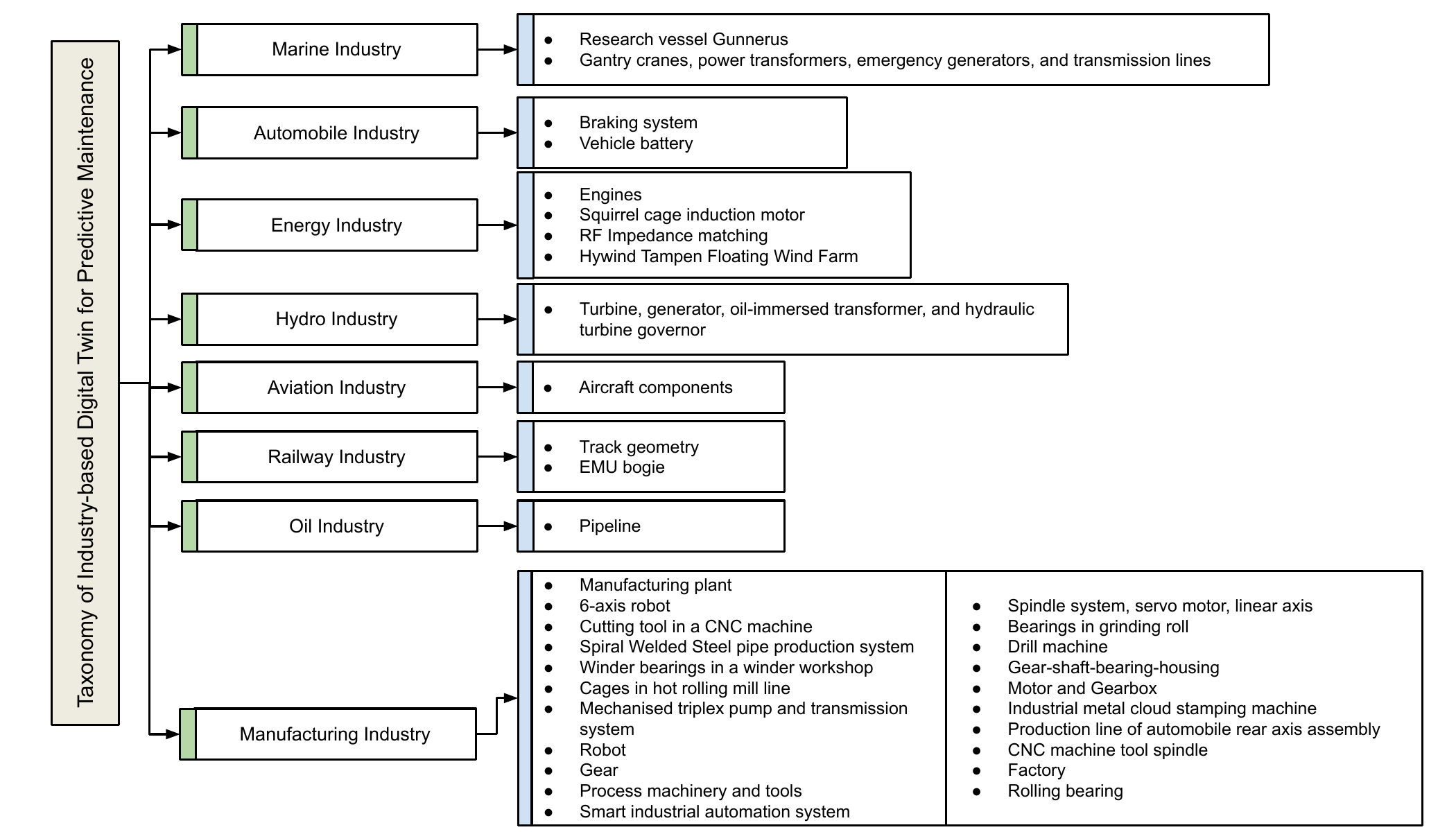}
    \caption{Taxonomy of industrial components using Digital Twin for Predictive Maintenance}
    \label{fig:taxonomyindustrydtpdm}
\end{figure*}

Cattaneo and Macchi \cite{72laura2019supportmachine} argued that historical run-to-failure data, which is essential for DT-based PdM, might not be available for the asset under investigation. Unlike other works that assume that this data might be available, \cite{72laura2019supportmachine} proposes a DT-based approach to simulate a drilling machine’s degradation process without strictly relying on the run-to-failure data at the beginning. The authors first compute a Root Mean Square (RMS) value for the machine’s acceleration signal to assess the amount of dissipated energy and then use a novelty detection approach \cite{73pimentel2014noveltydetection} to determine the confidence interval of RMS values for the machine operating in a healthy state. The confidence interval is then used to assess the health of the machine (i.e., healthy, abnormal, or faulty) and take necessary maintenance action. Similarly, Yu et al. \cite{74yu2023dynamicmodel} proposed a DT-based fault diagnosis approach without fault data. The authors developed a DT of gear-shaft-bearing-housing and used the model to simulate fault state data and healthy state data. The parameters for the DT model are optimized using the particle swarm optimization (PSO) algorithm. Furthermore, Selçuk et al. \cite{75Selcuk2021synthetic} proposed the use of DT to generate synthetic healthy and faulty conditions data for a motor and gearbox circuit. Pecora et al. \cite{76pecora2023monitoringmachine} proposed a DT-based approach that extracts information from the machine’s data to monitor the machine’s condition and detect performance degradation or anomalies. The authors used the CNN-BiLSTM approach to predict the current intensity and then detected anomalies using threshold-based Nelson rules. Wei et al. \cite{77wei2022augmented} proposed a DT and augmented reality-based PdM system for production lines. The proposed system uses historical data to determine the equipment health threshold and this threshold is then used in combination with real-time operational data to predict faults in the production line. Xue et al. \cite{78xue2022cncmachine} a DT-based framework to diagnose faults in CNC machine tools. Borangiu et al. \cite{79Borangiu2023robothealth} described DT-based layered architecture to monitor the health of industrial robots integrated into multi-resource manufacturing systems. The system uses a threshold-based approach which stops the robot if the real-time sensor collected data exceeds the predefined threshold value. Süve et al. \cite{80suve2021predictive} introduced an IoT and DT-based framework to predict the degradation and failures of machines in production environments. The authors demonstrated the effectiveness of the proposed approach using two use cases: 1) to predict failures of hydraulic accumulators and 2) to predict degradation of turbofan engines. Meng et al. \cite{81meng2023prediction} proposed a data/physics-based DT model for determining the remaining useful life of rolling bearings. The proposed approach uses a neural network to predict the useful life of rolling bearings. In the context of the oil industry, Lian et al. \cite{82lian2022application} proposed a DT-based approach to diagnose the health of crude oil pipelines. The approach uses DT to predict the transmission volume of the pipeline and compares the predicted volume with the actual volume to determine pipeline accidents such as leakage, blockage, or any other abnormalities. Figure \ref{fig:taxonomyindustrydtpdm} shows the taxonomy of the equipment or components used to create DT models for different industries in the literature.

In summary, the highest percentage, approximately $50\%$, of the implementation of DT for PdM in industrial engineering is attributed to the manufacturing industry, demonstrating a broad and diverse application of DT in various machinery and processes. This is followed by the energy industry, which accounts for around $15\%$ of implementations, particularly focusing on critical infrastructure such as wind farms and power transformers. In contrast, the least implementation is seen in the marine and railway industries. This suggests that further research and development are needed to increase the adoption of DT technology in other sectors, such as the logistics, transportation, vehicular, and aerial industries, where PdM could greatly enhance operational efficiency and reliability.

\section{Taxonomy of Digital Twin Algorithms for Predictive Maintenance in Industrial Engineering}
\label{sec:taxonomy}
\begin{figure}[htbp]
    \centering
    \includegraphics[width=\columnwidth]{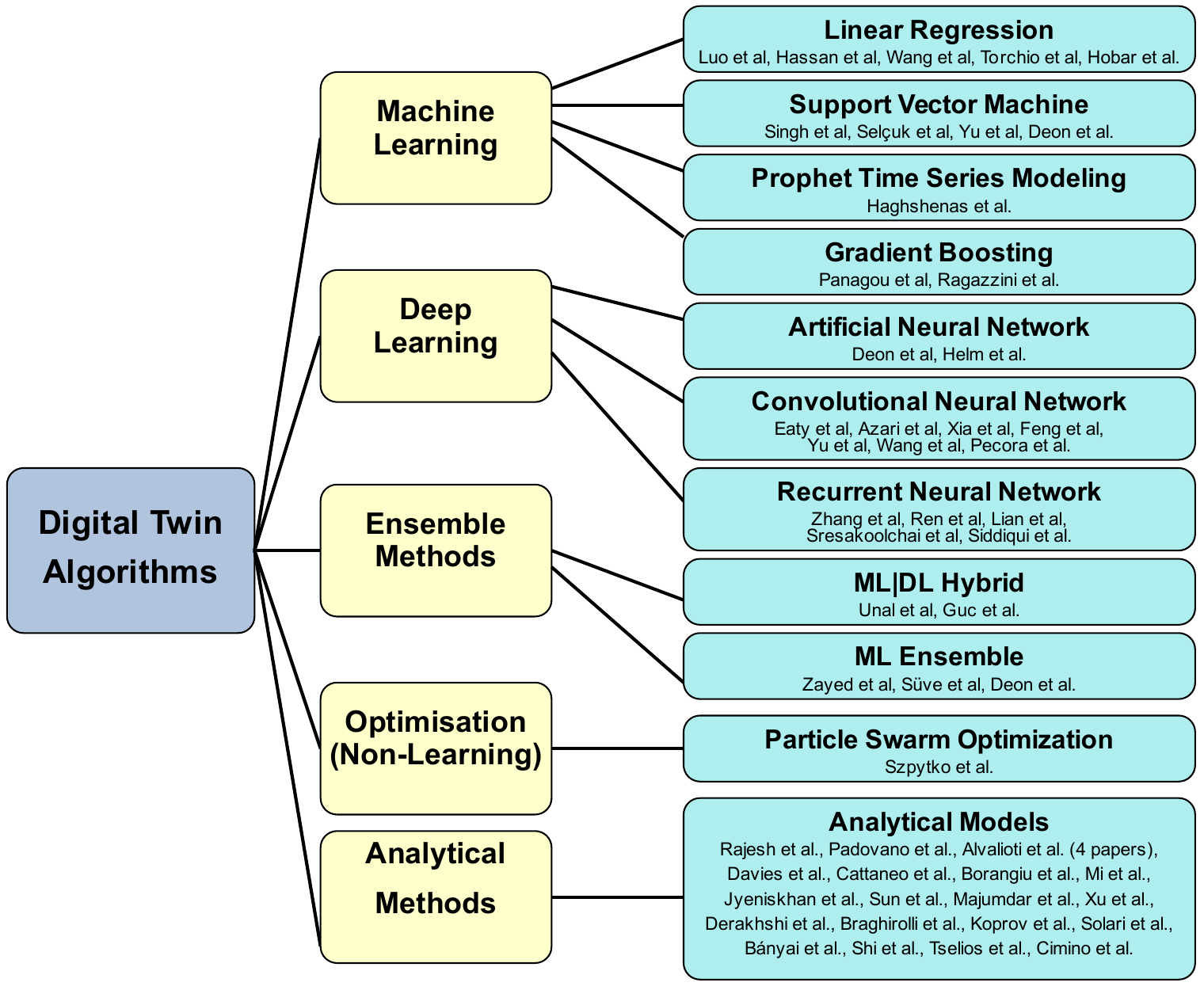}
    \caption{Taxonomy of Digital Twin algorithms for Predictive Maintenance}
    \label{fig:taxonomydtpdm}
\end{figure}

The classification of DT algorithms for PdM is organized into several key groups: Machine Learning, Deep Learning, ensemble methods, non-learning Optimization and analytical methods. The proposed taxonomy is presented in Figure \ref{fig:taxonomydtpdm}.
Different works use different machine learning and deep learning algorithms for PdM.
Zhang et al. \cite{43zhang2022digital} use the time series engine data collected using sensors and the event data registered by the ship operators to identify the normal operating, fault conditions, and run-to-failure data. They used the normal operating data to train a Long Short-Term Memory (LSTM)-based Variational Autoencoder (VAE) model to detect anomalies, i.e., deviations in the engine’s normal operating conditions. Furthermore, the authors used the run-to-failure data to predict the remaining useful life of the ship’s engine using an LSTM network. Szpytko and Duarte \cite{44szpytko2021digital} use the information regarding the historical degradation data of all the components in the system, previously planned processes, and system structures to enable entity managers to the container terminal to make faster and optimal scheduling decisions. The data is collected using Supervisory Control And Data Acquisition (SCADA) and Systems, Applications \& Products in Data Processing (SAP) systems. The authors used the Particle Swarm Optimization (PSO) approach to obtain a solution for the objective function that aims to minimize the gantry crane’s inefficiency. Eaty and Bagade \cite{46eaty2023digital} employ a battery dataset by the NASA prognostics center for Excellence to predict the state of charge using the Kalman filter. The state of health is predicted using the Alexnet model based on the incremental learning technique. Deon et al. \cite{47deon2022digital} compare neural networks, gradient boosting, random forest, SVM, self-organized maps, and k-means to predict failures related to fuel injection, cooling, air intake, exhaust, lubrication, and mechanical components of an engine. Singh et al. \cite{48singh2023building} apply the exponential degradation model to predict the remaining useful life of an induction motor and SVM model with Gaussian, cubic, fine Gaussian, linear, quadratic, medium, and coarse Gaussian kernels to detect faults in the motor. Guc and Chen \cite{49guc2022smart} run RFIM fault classification using the Ensemble Bagged Decision Tree model and Haghshenas et al. \cite{50haghshenas2023predictive} utilize the Prophet prediction algorithm for predicting bearing failure.

Wang et al. \cite{52wang2022digital} experiments with 1 Dimension convolutional Neural Network (1D-CNN) and Bidirectional Gated Recurrent Unit (BiGRU) algorithms to develop a transfer learning-based fault diagnoses model. A small sample of the dataset consisting of operating statuses of a thrust ball bearing is obtained from DT to initialize the transfer learning model. The real- time data from the physical system is then used to fine-tune the initial model and obtain a target model. The proposed transfer learning model is compared with Back Propagation Neural Network, K-Nearest Neighbour (kNN), Extreme Learning Machine, and Recurrent Radial Basis Function approaches. Heim et al. \cite{53heim2020predictive} used a neural network-based model for the prediction of the failure likelihood of aircraft components. Sresakoolchai and Kaewunruen \cite{54sresakoolchai2023railway} employ the Actor-Critic Reinforcement learning (RL) algorithm to perform maintenance activities. The selected maintenance activity improves the track geometry parameters and reduces the occurrence of track component defects. The track geometry and defects are the states. After taking action, the RL environment generates new states. Rewards are calculated in terms of maintenance costs and defect generates penalties.

Luo et al. \cite{58luo2020hybrid} examine linear regression, decision tree regression, random forest regression, and support vector regression to predict the remaining useful life of the cutting tool. Random forest regression gave the best result. The particle filtering algorithm is then used to revise the values of predicted states from the DT model using prediction from random forest. Unal et al. \cite{59unal2022data} used Random Forest, GBT, LSTM, SVR, KNN, and multi-layer perceptron on AML workshop data. Ren et al. \cite{60ren2023edge} proposed using LSTM for the prediction of the remaining useful life of winder bearings and IFCNN to diagnose faults in the bearings. Panagou et al. \cite{61panagou2022explorative} used XGBoost for fault detection in hot rolling mill cages. Zayed et al. \cite{62zayed2023efficient} used kNN, decision tree (CART), and random forest to predict fault predictions in triplex and transmission systems. CART outperforms other algorithms and FPA is the best for feature selection.

Feng et al. \cite{65feng2023digital} use CNN to predict the remaining useful life of a gear in a gearbox. Hassan et al. \cite{66hassan2024experience} adopt auto-regression with an exogenous variable model to predict degradation and component replacement. Panagou et al. \cite{68panagou2022feature} examine the XGBoost model to identify the most important sensors. Siddiqui et al. \cite{69siddiqui2023artificial} apply the NARX neural network to predict anomalies. Yu et al \cite{74yu2023dynamicmodel} utilize CNN to predict the state of a gear. The algorithm is compared with the Gaussian Mixture Model, AutoEncoder+K-means, support vector data description, SVM, random forest, LSTM, and MLP. Selçuk et al. \cite{75Selcuk2021synthetic} compared the performances of cubic SVM, ensemble bagged trees, naive Bayesian, weighted kNN, and quadratic discriminant methods for fault detection using publicly available and synthetic datasets. Ensemble bagged trees had the highest accuracy. Pecora et al. \cite{76pecora2023monitoringmachine} compare the performances of CNN-BiSTM and ARIMA for forecasting current intensity to supervise the machine’s condition. Wang et al. \cite{55wang2021complex} employ a grey forecasting model to predict fault in an EMU bogie. Lian et al. \cite{82lian2022application} compared the performances of LSTM and GRU algorithms to predict pipeline transmission volume to detect pipeline accidents. Süve et al. \cite{80suve2021predictive} compare Gaussian Naïve Bayes, Adaptive Random Forest, and LightGBM algorithms for identifying the failure of a hydraulic accumulator and predicting the degradation of a turbofan engine.

Works demonstrated in \cite{aivaliotis2021degradation, 45rajesh2019digital, 56padovano2021prescriptive, 57aivaliotis2019use, 63aivaliotis2019methodology, 64aivaliotis2023methodology, 67banyai2022real, 70davies2022digital, 71mi2021prediction, 72laura2019supportmachine, 79Borangiu2023robothealth} does not include a learning algorithm for PdM. P.K. et al. \cite{45rajesh2019digital} simulated the wear of brake pads at different vehicle speeds to identify the remaining useful life using an analytical method. Padovano et al. \cite{56padovano2021prescriptive} simulated different scenarios to evaluate the overdue delivery of an order with and without implementing a DT-based prescriptive maintenance system. Aivaliotis et al. \cite{57aivaliotis2019use} examine the output from a real machine and simulated output from the DT model to compute the remaining useful life of a machine. Similarly, Aivaliotis et al. \cite{aivaliotis2021degradation}, Aivaliotis et al. \cite{63aivaliotis2019methodology}, and Aivaliotis et al. \cite{64aivaliotis2023methodology} apply the DT model to simulate scenarios for remaining useful life prediction. Bányai and Bányai \cite{67banyai2022real} utilize a DT-based optimization approach to identify optimal maintenance policy for the management of energy efficiency and less GHG emissions. Davies et al., while \cite{70davies2022digital} examine a mathematical formula to calculate the remaining useful life of the spindle system, servo motor, and linear axis. Cattaneo and Macchi \cite{72laura2019supportmachine} use acceleration information from the drill machine rotary equipment while the machine is in an operational state (determined based on the PLC signal) to compute the RMS value for each component. RMS value is used to determine the state of the machine. Borangiu et al. \cite{79Borangiu2023robothealth} utilize a predefined threshold to predict faults in industrial robots. A summary of these contributions is provided in Table \ref{tab:table5_a}, \ref{tab:table5_b}.

\begin{table*}[!t]
    \centering
    \renewcommand{\arraystretch}{1.2}
    \begin{threeparttable}
        \caption{Applications of Digital Twin in Predictive Maintenance}
        \label{tab:table5_a}
        \begin{tabular}{p{0.03\textwidth} p{0.08\textwidth} p{0.05\textwidth} p{0.35\textwidth} p{0.26\textwidth} p{0.08\textwidth}}
            \toprule
            \textbf{Ref} & \textbf{Authors} & \textbf{Year} & \textbf{Application Area} & \textbf{Algorithm} & \textbf{Formula Mentioned} \\
            \midrule
            \cite{45rajesh2019digital} & P.K. et al. & 2019 & Digital Twin for Predictive Maintenance of Automotive Brake Pads & NR & No \\
            \cite{57aivaliotis2019use} & Aivaliotis et al. & 2019 & Digital Twin for predictive maintenance in manufacturing resources & Simulation / Predefined threshold & No \\
            \cite{63aivaliotis2019methodology} & Aivaliotis et al. & 2019 & Methodology for enabling Digital Twin using advanced physics-based modeling in predictive maintenance (Industrial robot) & NR & No \\
            \cite{72laura2019supportmachine} & Cattaneo and Macchi & 2019 & Machine prognostics in industrial drilling operations. & Exponential Degradation Model (EDM) & Yes \\
            \cite{53heim2020predictive} & Heim et al. & 2020 & Aircraft Component Failure Prediction & Neural network-based model & No \\
            \cite{58luo2020hybrid} & Luo et al. & 2020 & Remaining useful life prediction of cutting tool & Linear regression, Decision tree regression, Random Forest regression, Support vector regression, Particle filtering & Yes \\
            \cite{44szpytko2021digital} & Szpytko and Duarte & 2021 & Scheduling Optimization in Container Terminals & Particle Swarm Optimization (PSO) & Yes \\
            \cite{75Selcuk2021synthetic} & Selçuk et al. & 2021 & Predictive maintenance for industrial equipment using vibration data analysis. & Cubic SVM, Ensemble bagged trees, Naïve Bayesian, Weighted kNN, Quadratic discriminant & Yes \\
            \cite{55wang2021complex} & Wang et al. & 2021 & Fault prediction in an EMU bogie & Grey forecasting model & Yes \\
            \cite{aivaliotis2021degradation} & Aivaliotis et al. & 2021 & Predictive maintenance of industrial robots through degradation curves integration in physics-based models & Physics-based models, Degradation Curves & Yes \\
            \cite{56padovano2021prescriptive} & Padovano et al. & 2021 & Evaluation of overdue order delivery with prescriptive maintenance & NR & No \\
            \cite{71mi2021prediction} & Mi et al. & 2021 & Decision-making approach for predictive maintenance supported by Digital Twin & NSGA-II Hybrid Algorithm, Mathematical Programming, Monte-Carlo Random Simulation, BP Neural Network & Yes \\
            \cite{47deon2022digital} & Deon et al. & 2022 & Engine Failure Prediction & Artificial Neural networks, Gradient Boosting, Random Forest, SVM, Self-organized maps, k-means & Yes \\
            \cite{49guc2022smart} & Guc and Chen & 2022 & Fault Classification for RFIM & Ensemble Bagged Decision Tree & No \\
            \cite{33thakur2023effective} & Wang et al. & 2022 & Transfer Learning-based Fault Diagnosis for Thrust Ball Bearing & 1D-CNN, BiGRU, Transfer Learning & Yes \\
            \cite{59unal2022data} & Unal et al. & 2022 & Predictive maintenance and operational efficiency in industrial process machinery. & Random Forest, GBT, LSTM, SVR, KNN, Multi-layer perceptron & No \\
            \cite{61panagou2022explorative} & Panagou et al. & 2022 & Fault detection in hot rolling mill cages & XGBoost & No \\
            \cite{68panagou2022feature} & Panagou et al. & 2022 & Sensor importance identification & XGBoost & No \\
            \cite{76pecora2023monitoringmachine} & Pecora et al. & 2022 & Forecasting current intensity for machine supervision & CNN-BiSTM, ARIMA & No \\
            \cite{82lian2022application} & Lian et al. & 2022 & Pipeline accident detection & LSTM, GRU, PCA & Yes \\
            \cite{80suve2021predictive} & Süve et al. & 2022 & Failure identification of a hydraulic accumulator & Gaussian Naïve Bayesian, Adaptive Random Forest, LightGBM & No \\
            \cite{67banyai2022real} & Bányai and Bányai & 2022 & Optimal maintenance policy for energy management & Markov Decision Process, Howard's Policy Iteration Technique, Evolutionary Heuristics & Yes \\
            \cite{70davies2022digital} & Davies et al. & 2022 & Remaining useful life prediction of spindle system, servo motor, and linear axis & Mathematical algorithm for RUL & Yes \\
            \cite{24luo2020hybrid} & Zhang et al. & 2023 & Predictive Maintenance for Ship Engines & LSTM-based Variational Autoencoder (VAE) & Yes \\
            \cite{46eaty2023digital} & Eaty and Bagade & 2023 & Battery Health Prediction & Kalman filter, AlexNet model with incremental learning & Yes \\
            \cite{48singh2023building} & Singh et al. & 2023 & Induction Motor Remaining Useful Life Prediction & Exponential degradation model, SVM with various kernels & Yes \\
            \cite{50haghshenas2023predictive} & Haghshenas et al. & 2023 & Bearing Failure Prediction & Prophet prediction algorithm & Yes \\
                        \bottomrule
        \end{tabular}
    \end{threeparttable}
\end{table*}
\begin{table*}[!t]
    \centering
    \renewcommand{\arraystretch}{1.2}
    \begin{threeparttable}
        \caption{Applications of Digital Twin in Predictive Maintenance}
        \label{tab:table5_b}
        \begin{tabular}{p{0.03\textwidth} p{0.08\textwidth} p{0.05\textwidth} p{0.35\textwidth} p{0.26\textwidth} p{0.08\textwidth}}
            \toprule
            \textbf{Ref} & \textbf{Authors} & \textbf{Year} & \textbf{Application Area} & \textbf{Algorithm} & \textbf{Formula Mentioned} \\
            \midrule
            \cite{54sresakoolchai2023railway} & Sresakoolchai and Kaewunruen & 2023 & Track Maintenance Optimization & Advantage Actor-Critic Reinforcement Learning & Yes \\
            \cite{60ren2023edge} & Ren et al. & 2023 & Prediction and fault diagnosis in winder bearings & LSTM, IFCNN & No \\
            \cite{62zayed2023efficient} & Zayed et al. & 2023 & Fault prediction in triplex and transmission systems & Hybrid optimization techniques (FPA, PSO, HHO, JA, GWO, SSA) with ML models (KNN, CART, RF) & Yes \\
            \cite{65feng2023digital} & Feng et al. & 2023 & Remaining useful life prediction of a gear in a gearbox & CNN & Yes \\
            \cite{69siddiqui2023artificial} & Siddiqui et al. & 2023 & Anomaly prediction using NARX neural network & NARX neural network & Yes \\
            \cite{74yu2023dynamicmodel} & Yu et al. & 2023 & Gear state prediction & CNN, Gaussian Mixture Model, AutoEncoder+K-means, Support vector data description, SVM, Random Forest, LSTM, MLP & Yes \\
            \cite{64aivaliotis2023methodology} & Aivaliotis et al. & 2023 & Methodology for enabling dynamic Digital Twin & Nonlinear Least Squares method & Yes \\
            \cite{79Borangiu2023robothealth} & Borangiu et al. & 2023 & Fault prediction in industrial robots & Simulation / Predefined threshold & No \\
            \cite{66hassan2024experience} & Hassan et al. & 2024 & Degradation and component replacement prediction & Autoregression with exogenous variable model & No \\
            \bottomrule
        \end{tabular}
    \end{threeparttable}
\end{table*}

\section{Case Study}
\label{sec:case_study}
Based on the results of this systematic review and our vision of a layered architecture for DT-PdM, we create a prototype system along with an application in a smart aviation industry scenario, by leveraging IIoT, edge, and cloud computing systems \cite{ismail2015implementation}. Certainly, our proposed architecture can be first evaluated for various scenarios via modeling and simulation using one of the tools described in Table  \ref{tab:software}. It connects the IIoT devices and sensors with the edge and cloud computing infrastructure.

\subsection{A Smart Aviation Application with Integrated Edge and Cloud Computing Environment}
\begin{figure*}[!ht]
    \centering
    \includegraphics[width=0.8\textwidth]{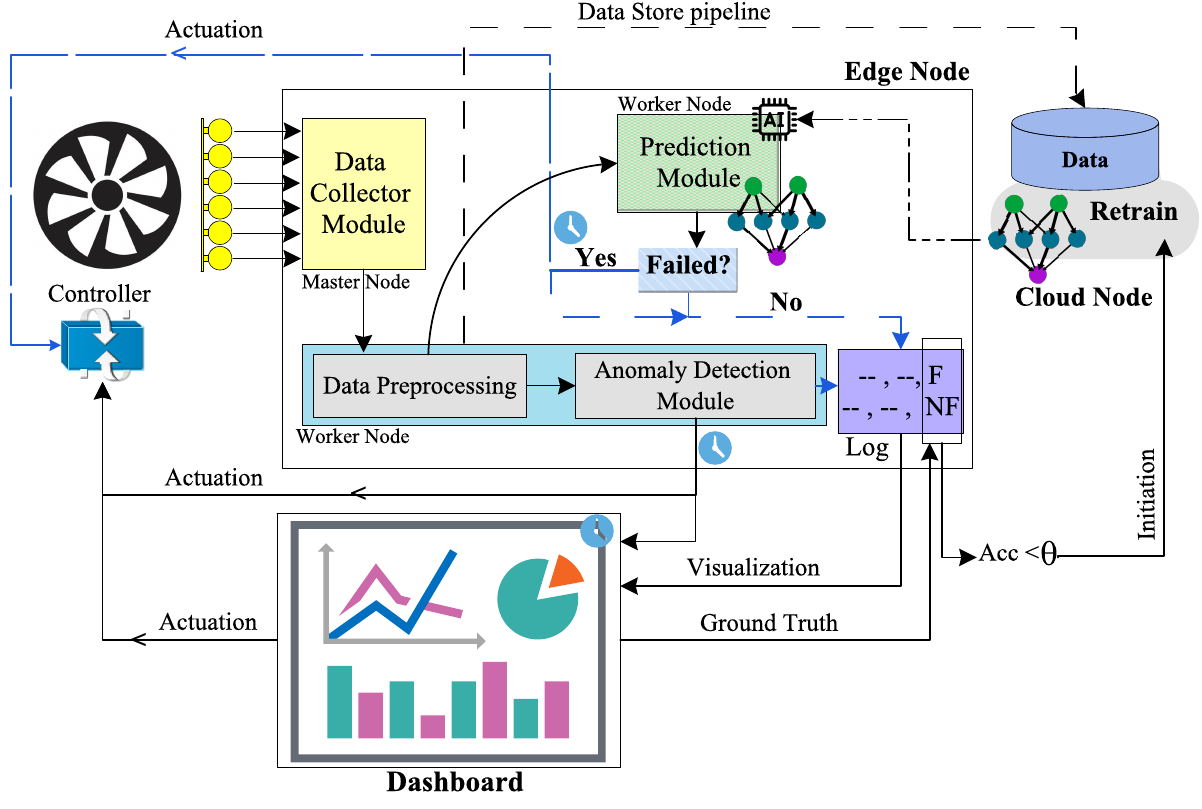}
    \caption{Airplane turbine failure prediction architecture}
    \label{fig:casestudy}
\end{figure*}

We will now discuss a case study of leveraging the edge and cloud computing integrated systems for IIoT applications, via an application demonstrator of DT-PdM in the aviation industry. The network topology that we use is based on the master-worker model \cite{ismail2008formal, ismail2012modeling}. It connects different IoT devices and sensors with edge and cloud infrastructure to process data and tasks on worker nodes in the edge and cloud environment. Figure \ref{fig:casestudy} represents our case study for airplane turbine engine monitoring and failure prediction using a prototype framework that we developed. It shows how interconnected components seamlessly cooperate for decision-making, anomaly detection, and proactive maintenance.

In this IIoT application, IIoT sensors for engine state monitoring are used to sense the engine condition in terms of temperature, pressure, vibration, acoustics, fuel flow, and energy consumption, and send the data to the edge computing node for pre-processing and failure prediction of the engine. The edge node sends the pre-processed data received from the IIoT sensors to the Cloud node for storage. The prediction model on the edge computing node is the result of machine learning training in the Cloud. Training the pre-processed dataset in the Cloud is achieved offline so that the development of the prediction model does not impact the prediction performance on the Edge. The prediction model on the Edge is updated whenever a new model is produced by the Cloud. Our framework for failure prediction uses the following hardware and software components, as described below.

The  PdM architecture in three interconnected layers: the Data Collection and Preprocessing Layer along with the Prediction and Anomaly Detection Layer (Edge Node), the Retraining and Data Storage Layer (Cloud Node), and the Dashboard or Visualization deck. Each layer plays a distinct yet collaborative role, facilitating a robust PdM system capable of real‑time intervention and continuous improvement through iterative feedback loops.
\subsection{Hardware Components}
\subsubsection{IIoT Sensors}  
We embed industrial-grade IoT sensors into machine components to capture vibration, acoustic signals, temperature, humidity, pressure and energy-consumption metrics at regular intervals. Each sensor samples and buffers its readings locally, performs a quick check (e.g. range validation), while preprocess it and sends the data over a secure link to the Master Node. If connectivity drops, the sensor caches measurements and transmits them as soon as the network restores.
\subsubsection{Edge Node}
We deploy compact edge servers—our laptop with the Ubuntu operating system, rtx 3070 4GB with 317MB usable space, 16GB RAM or typically NVIDIA Jetson Nano or AGX modules—within one or two network hops of the sensors. These devices perform real-time data cleaning, normalization, and lightweight anomaly detection using on-board GPUs. When they detect an outlier or failure signature, they issue an immediate alert packet to the Master Node and log full event details for later analysis. They also buffer incoming streams during upstream outages and automatically resume upload. While facing incoming failures, the node periodically update the prediction model from cloud node.
\subsubsection{Cloud Node}  
We run our Cloud Node on a high-performance cluster equipped with two NVIDIA A5000 GPUs and 4TB Hard disk, 64GB RAM, Ubuntu operating system. Here, we aggregate all sensor and edge-processed data into a normalized central time series database and retrain our predictive maintenance models on the full historical dataset. Once we validate a new model based on present failures, we push the updated weights back to each Edge Node for continuous, low-latency inference.
\subsection{Software Components}
\subsubsection{Master Node}
We operate gateway software on the Master Node to authenticate devices, decrypt incoming packets, and tag every record with a unified timestamp. The gateway normalizes diverse protocols (e.g. MQTT, OPC-UA, Modbus) into a single data stream, balances load across Worker Nodes, and monitors throughput in real time. In this node, we use predefined scripts for copying the contents to the cloud node while keeping the connection steady and passing to the worker node for further processing.
\subsubsection{Worker Node}
On each Worker Node, we run containerized services to preprocess incoming streams—filling small gaps and filtering noise—before applying a fault-prediction routine (for example, a lightweight Artificial Neural Network (ANN)). If the routine flags a potential failure, the Worker Node sends control commands directly to the PLCs to trigger corrective actions. We log every inference result along with operator feedback to our Cloud database so we can measure accuracy and refine our models during the next retraining cycle.
\subsection{Experimental Validation}
\subsubsection{Experimental Setup and Environment}  
We run all training and evaluation on a cloud server with an NVIDIA RTX A5000 GPU, 64 GB RAM, Ubuntu, and Python 3.10. We installed PyTorch 2.0.1, Scikit-learn 1.2.2, Pandas 2.0.3, NumPy 1.24.4, Matplotlib 3.7.1, and Imbalanced-learn 0.10.1. To mirror real-world constraints, we replicate the Edge Node setup on a smaller device. The Cloud and edge devices are part of our Intelligent Distributed Computing and Systems (INDUCE) Lab of the United Arab Emirates University.
\subsubsection{Dataset Acquisition and Description}  
We use a publicly available dataset \cite{kaggledataset} which contains sensor readings, such as vibration, acoustic signals, temperature, humidity, pressure, and energy consumption metrics, and a binary 'Fault' label that marks each record as normal or faulty. We pre-process the data using minmax and Powertransformer \cite{taha2021novel}-based data cleaning and normalization strategies for training and validation. In addition, we used SMOTE \cite{chawla2002smote} to balance the classes before training.
\subsubsection{Deep Learning Model Architecture}
We build our failure detector similar to \cite{dworakowski2017artificial} with an encoder of four dense layers (512, 512, 256, 128 units) using LeakyReLU activations. We add a residual connection \cite{he2016deep} to carry forward the input features and then stacked two 64-unit dense layers before the final Softmax binary output. This design keeps the model small enough for Edge deployment while retaining enough capacity to learn complex fault patterns.
\subsubsection{Training Methodology}  
We employ stratified 5-fold cross-validation, apply SMOTE \cite{chawla2002smote} only on each training fold to avoid information leakage. We optimize with AdamW and schedule learning rates using OneCycleLR. To focus the model on rare fault cases, we adopt focal loss, and we also apply mix-up augmentation \cite{onishi2023rethinking} to improve generalization.
\begin{figure*}[!ht]
    \centering
    \includegraphics[width=\textwidth]{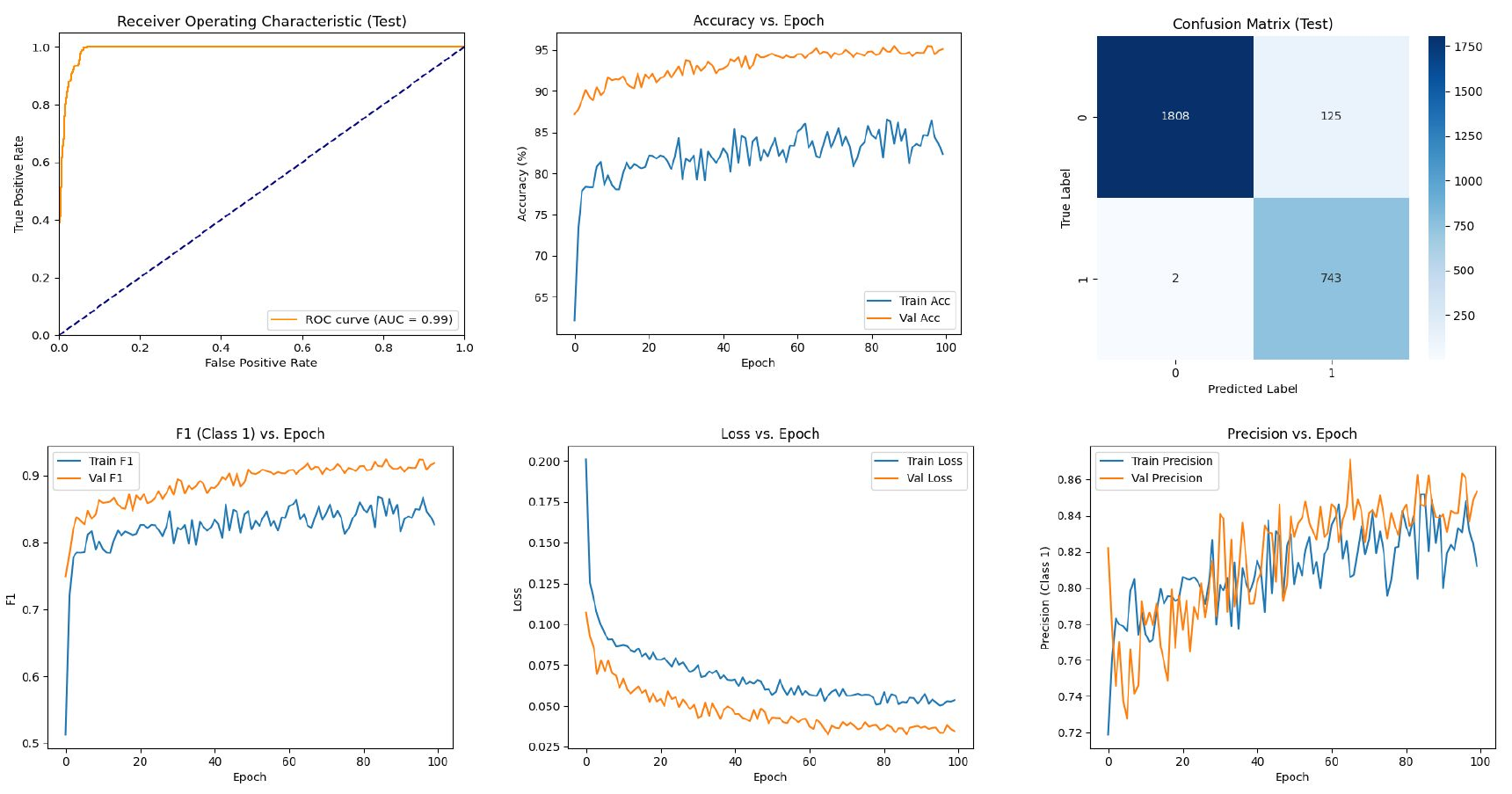}
    \caption{Performance Evaluation of airplane turbine fault detection model. Six metrics show model evaluation: (a) ROC curve with AUC=0.99; (b) Accuracy vs. Epoch reaching 95\% validation accuracy; (c) Confusion matrix showing 1808 true negatives, 743 true positives, 125 false positives, and only 2 false negatives; (d) F1-score for fault class exceeding 0.9; (e) Consistent loss decrease demonstrating proper convergence; (f) Precision improvement stabilizing at approximately 0.85. The extremely low false negative rate (2/745) highlights the effectiveness of the model for PdM applications.}
    \label{fig:casestudy_exp}
\end{figure*}
\subsubsection{Numerical Results Analysis}
Figure~\ref{fig:casestudy_exp} summarizes our performance \cite{zhou2021machine}. The ROC curve in panel (a) reaches an AUC of 0.99, showing excellent discrimination between normal and faulty cases. The accuracy plot in (b) climbs to 95 percent on the validation set, and the loss curve in (e) drops smoothly, indicating steady learning without overfitting. In the confusion matrix (panel c), we correctly identify 743 failures and 1 808 normal states, while producing only 125 false alarms and missing just 2 failures. The precision curve in (f) stabilizes around 0.85 and the F1 score in (d) stays above 0.90, confirming reliable detection. These results demonstrate that our model can run in real time on Edge hardware and still catch almost all fault events, making it well-suited for predictive maintenance.
\section{Challenges and Research Direction}
Digital Twin (DT) involves creating a virtual model of physical systems and has gained significant attention as a tool for Predictive Maintenance (PdM) in various industries. By leveraging real-time data, DTs emerge for the deployment of Smart Factories. In particular, for PdM, DTs can predict system failures, optimize maintenance schedules, and enhance the overall efficiency of industrial operations. However, the implementation of DT for PdM presents several challenges that require focused research to overcome. This section highlights the key challenges for efficient, real-time, and sustainable DT for PdM, and explores potential opportunities for future research.

\subsection{High-Fidelity Modeling and Real-Time Data Integration:}
\begin{itemize}
    \item \textbf{Challenge: } One of the key challenges in DT implementation is achieving high-fidelity modeling that accurately mirrors the physical system's behavior. The DT must integrate real-time data to provide precise predictions, but the accuracy of these models can be compromised due to discrepancies between the virtual and physical components. For instance, inaccuracies in sensor data or delays in data transmission can lead to errors in the DT model, thereby reducing its effectiveness in PdM \cite{yang2022digital}.
    \item \textbf{Future Research Direction: } To address this, there is a need for research into advanced simulation techniques, including physics-based modeling, machine learning (ML), and hybrid approaches that combine these methodologies. Additionally, real-time calibration of models using live data from sensors could improve the accuracy and reliability of DTs. Exploring the integration of quantum computing in simulation processes could also be a promising direction for achieving higher accuracy in complex systems. Recent studies have emphasized the importance of integrating real-time data synchronization and analytics between physical devices and their digital counterparts for precise monitoring and management \cite{turn0search5}. In addition, the development of high-fidelity DT models has been shown to enable real-time monitoring, fault diagnosis, and performance evaluation of physical entities \cite{turn0search29}.
\end{itemize}

\subsection{Data Management and Analytics}
\begin{itemize}
    \item \textbf{Challenge: } The vast amount of data generated by sensors in a DT for PdM framework poses significant challenges in data management and analytics. Efficient transmission, pre-processing, and analytics of this data are crucial for accurate predictions and maintenance decisions \cite{ismail2015implementation}. The current state-of-the-art in data analytics still faces difficulties in managing heterogeneous data sources and ensuring data quality, which can lead to suboptimal maintenance strategies \cite{liu2022digital}.
    \item \textbf{Future Research Direction: } Development of advanced, scalable data fusion and analytics frameworks that can efficiently handle the vast and heterogeneous data generated by sensors is essential. This research should focus on creating robust architectures capable of real-time processing and integration of multi-source data, including structured and unstructured data from various sensor types, legacy systems, and external sources. Leveraging edge computing and distributed data processing techniques could reduce latency and improve data handling at the point of generation, while cloud computing could be utilized for large-scale storage and deeper analytics, and edge computing, closer to the IIoT devices, can be used for fast decision-making and inference \cite{materwala2023qos}. Another key research avenue is the application of machine learning and AI to enhance data quality and analytics, enabling the detection and correction of data inconsistencies, missing values, and noise. By developing algorithms that can adaptively learn from data in real-time, DT systems could generate more accurate predictions, leading to more effective and timely maintenance strategies. The role of data fusion in the DT ecosystem for predictive maintenance has been highlighted as a critical component in recent research \cite{turn0search28}. In addition, the integration of AI techniques has been shown to significantly enhance the performance of DTs in industrial applications \cite{turn0search16}.
\end{itemize}

\subsection{Unified Development Framework and Tools:}
\begin{itemize}
    \item \textbf{Challenge: } While several researchers have explored the framework of PdM in DT, the existing models are either overly generic, lacking detailed hierarchical structures, or are focused solely on specific applications. Current research is notably deficient in offering a standardized framework for the implementation of DT. Key issues, such as determining the optimal timing for DT implementation, ensuring compliance with existing standards, and selecting appropriate modeling methods for diverse machinery, remain unresolved in both academic and industrial contexts.
    \item \textbf{Future Research Direction: } Addressing these challenges necessitates the development of a standardized framework for PdM, DT, encompassing system architecture, workflow processes, and modeling techniques. To provide clear and actionable guidance for practical applications, further research efforts should prioritize the establishment of a robust and standardized DT framework. A requirement-based roadmap for standardized predictive maintenance automation using DT technologies has been proposed to support this requirement \cite{turn0academia51}. Moreover, the development of distributed DT frameworks has been identified as a means to improve predictive maintenance in manufacturing assets \cite{turn0search7}.
\end{itemize}

\subsection{Cybersecurity and Privacy Concerns}
\begin{itemize}
    \item \textbf{Challenge: } Regarding the security and privacy concerns of DT for PdM systems, these technologies collect vast amounts of data to create their virtual models. However, similar to other information systems, they face cyber threats and attacks. The reliability and accuracy of DT models depend heavily on the integrity of data sourced from the physical world. If this data is compromised or attacked, it can lead to flawed models and analysis, resulting in incorrect simulations and decisions. In the worst cases, data theft or manipulation can occur, severely impacting the trustworthiness of the entire DT system. In addition, the collection and processing of vast amounts of data raise privacy concerns, particularly in industries that handle sensitive information, thus posing significant legal and ethical risks.
    \item \textbf{Future Research Direction: } Addressing cybersecurity and privacy concerns requires a multi-faceted approach. Research should explore the integration of advanced encryption techniques, secure communication protocols, and blockchain technology to enhance the security of DT systems \cite{ismail2019review}, similar to the results by \cite{hennebelle2024secure} in developing a blockchain-based secure and privacy-preserving system for IoT applications. Other privacy-preserving machine learning methods, such as federated learning, could be investigated to allow predictive models to be trained on decentralized data sources without compromising privacy. Recent studies have investigated the role of AI in providing cybersecurity for digital twin versions of various industries, highlighting the potential of AI in enhancing security measures \cite{turn0search14}. In addition, the application of DT in cybersecurity has been explored, emphasizing their role in reducing attacks in an increasingly connected world \cite{turn0search23}.
\end{itemize}

\subsection{Environmental Coupling Technologies}
\begin{itemize}
    \item \textbf{Challenge: } Currently, many DT for PdM implementations do not fully integrate the influence of external environmental factors into their virtual models. The interaction between physical assets and their surrounding environment, such as temperature, humidity, and operational stressors, is well-documented in real-world studies. However, translating these interactions into digital representations within the DT remains a significant challenge. Developing effective methods to digitally express these environmental interactions is essential for achieving more accurate and efficient predictions in future DT applications.
    \item \textbf{Future Research Direction: } Research should focus on creating adaptable environmental coupling models that can account for varying operational conditions across different industries and environments. ML could further enhance the predictive capabilities by identifying patterns and correlations between environmental factors and asset degradation. By creating standardized models and algorithms that are both flexible and context-specific, future DT systems could provide highly accurate predictions, allowing for more proactive maintenance strategies that account for both internal and external influences on asset performance. Recent studies have emphasized the importance of integrating environmental data into DTs to strengthen climate adaptation and mitigation decision-making \cite{turn0search6}. In addition, the development of distributed DT frameworks has been identified as a means to improve predictive maintenance in manufacturing assets \cite{turn0search7}.
\end{itemize}

\section{Discussion}
 There has been an increased interest in the deployment of digital twins in industrial engineering, thanks to the emergence of IoT, AI, big data analytics, federated and distributed learning, edge and cloud computing, and blockchain \cite{ismail2022artificial, ismail2014fsbd}. AI-DT-enabled PdM improves the accuracy of fault detection and life-prediction of industrial assets by combining the powers of real-time data acquisition, machine learning analytics, and robust simulation models. The evolution of PdM from basic simulation tools in 1991 to  DT systems powered by IoT underscores a path of innovation that will transform preventive maintenance paradigms into a proactive and cost-effective approach.
A seamless communication between the digital representation, the physical asset, and the infrastructure distributed ecosystems, is essential for the efficiency of AI-DT-enabled PdM, which we call DT2X, as shown in Figure \ref{fig:overviewdtpdm}, notably presenting an edge-based predictive maintenance architecture that has the potential to reduce unscheduled downtime and maintenance expenses. Our examination of software tools and architectural frameworks (Sections \ref{sec:overview}, \ref{sec:tools}), and requirements of DT for PdM (Section \ref{sec:requirements}, highlights that a DT must have high-fidelity data modeling, scalable, QoS-enabled and secure communication protocols and processing, for effective DT deployment. This should be underpinned by many methods used for anomaly detection, defect diagnosis, and usable life prediction are further demonstrated by the thorough taxonomy of AI algorithms provided in Section \ref{sec:taxonomy}. Despite the great potential of these approaches, the disparities in integration techniques and performance indicators among various industrial applications underscore the need for a uniform assessment framework.  This analysis shows that despite some approaches performing well in controlled settings, their scalability in practical situations is still unknown. The practical applications of these findings highlight the revolutionary potential of DTs in maximizing asset longevity and operating efficiency.  However, our survey also highlights some of the more general issues that our assessment pointed out, highlighting the need for more effort to create cohesive frameworks that can transcend the present constraints.
While working on such insights, we paint a comprehensive picture of the current landscape: DT-enabled predictive maintenance holds potential to revolutionize industrial operations, yet its success is contingent upon addressing significant technical and operational challenges that we highlighted in this paper. In this work, we bridge theoretical advances and practical applications, and establish a connection to future research directions.

\section{Conclusions and Future Works}
In this paper, we traced the evolution from traditional reactive and preventive maintenance to modern data-driven predictive approaches, demonstrated how real-time sensor data, advanced analytics, and robust simulation models are converging within DT frameworks to revolutionize asset management \cite{aivaliotis2021degradation,rojek2023artificial}. Our systematic review covers the state of the art, including detailed taxonomies of AI techniques, integration requirements, and practical case studies, which collectively illustrate the potential of PdM enabled by AI-powered DT to reduce unplanned downtime and optimize maintenance operations.

Despite some advancements in DT for PdM, several challenges persist that provide a fertile ground for future research. The development of standardized evaluation frameworks is essential to uniformly assess the performance of diverse AI algorithms across multiple industrial contexts \cite{smith2024integration}. In addition, more research is needed to enhance cybersecurity measures, particularly for real-time data exchange and remote monitoring on DT platforms \cite{doe2023cybersecurity}. Integration of SOTA systems with modern IoT architectures remains a complex task, which requires scalable solutions that seamlessly accommodate both new and old infrastructures. 

Such grounds can be further adapted to the advancement of data and data-driven analytics to improve the predictive accuracy and reliability of DT systems, as enhanced data management frameworks can better handle the inherent diversity and volume of industrial data \cite{liu2023data}. Moreover, there is a pressing need for a unified development framework that standardizes tools and methodologies across DT applications, facilitating smoother integration and interoperability \cite{robinson2023framework}. Finally, investigating environmental coupling technologies can offer insights into how external factors such as climate variations affect digital twin performance, thereby enabling more robust predictive models \cite{kim2024environmental}. Addressing these directions will not only bridge existing gaps in DT-enabled predictive maintenance practices, but will also set the stage for more efficient industrial systems. Overall, our review provides a solid foundation that both captures the current state of research and demarcates clear directions for future innovations in this dynamically evolving field.

\section{Acknowledgment}
This work was supported by the Emirates Center for Mobility Research, United Arab Emirates University.

\begin{IEEEbiography}[{\includegraphics[width=1in,height=1.25in, clip,keepaspectratio]{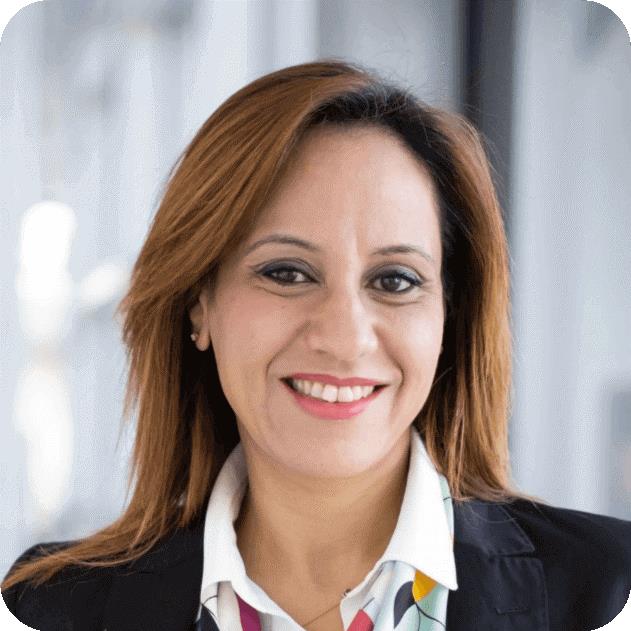}}]{Leila Ismail} is the Founding Director of the Intelligent Distributed Computing and Systems (INDUCE) Lab, an Associate Professor at the Department of Computer Science and Software Engineering of the College of Information Technology at the United Arab Emirates University (UAEU), and a Visiting Associate Professor in the Cloud Computing and Distributed Systems (CLOUDS) Lab, School of Computing and Information Systems, Faculty of Engineering and Information Technology at The University of Melbourne, Australia for the academic year 2022- 2023. She received her Ph.D. from the Grenoble Institute of Technology and French Institute for Research in Computer Science and Technology (INRIA) in France with a very honorable degree. She has vast industrial and academic experience at the Sun Microsystems Research and Development Center, France, on the design and implementation of intelligent systems, participated in a U.S. patent, and has been a lecturer at Grenoble Institute of Technology, France, an Adjunct Professor at the Digital Ecosystems and Business Intelligence Institute at Curtin University, Australia. She has been very active in creating smart efficient, and green IoT digital ecosystems responding to modern emergency needs in sustainable Smart Cities, enabled by IoT, AI, machine learning, deep learning, Big Data, and blockchain for different smart city applications domains, such as energy savings, smart transportation systems, and smart digital healthcare. She is active in international collaborations and publishes her research results in prestigious journals and international conferences. She served as Associate Editor of the International Journal of Parallel, Emergent, and Distributed Systems for several years, served as Chair, Co-Chair, and Track Chair for many IEEE international conferences, including being a General Co-Chair for IEEE DEST 2009, and a General Chair, Technical Program Chair, and Organizing Committee Chair for the 11th International Conference on Innovations in Information Technology 2015 (IIT’15) for which Dr. Ismail got the support of the IEEE Computer Society (HQs, USA) technical sponsorship. She is the Editor of Information Innovation Technology in Smart Cities, published by Nature Springer. She is widely invited as a keynote speaker to several conferences, including the Women in Data Science International Conference (WiDS 2021), organized by Stanford University, USA. She is the recipient of several awards and appreciation certificates. She is in the top world’s 2\% cited scientists in 2023 and 2024 in AI, Networking and Communication, Information and Communication Technologies by Elsevier and Stanford University, USA.
\end{IEEEbiography}
\begin{IEEEbiography}[{\includegraphics[width=1in,height=1.25in, clip,keepaspectratio]{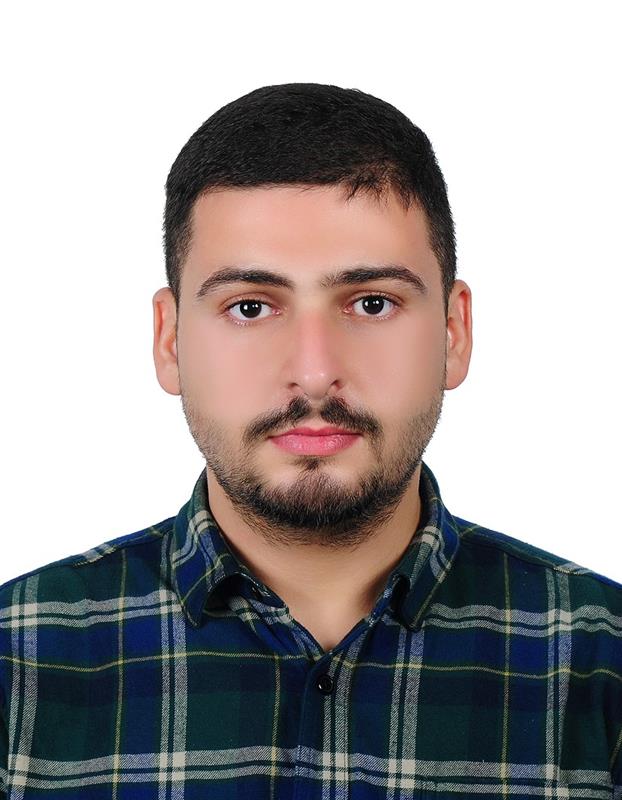}}]{Abdelmoneim Mohamed Abdelmoneim}  holds a Bachelor's degree in Architectural Engineering from Abu Dhabi University and a Master's degree from the Department of Architectural Engineering of the United Arab Emirates University (UAEU). His research focuses on sustainable developments, particularly Digital Twin-enabled asset management in Architectural Engineering and Construction Facility Management, and energy-efficient building design. He currently works as an architect in the industry, having previously served as a Research Assistant in the Department of Architectural Engineering at UAEU.
\end{IEEEbiography}
\begin{IEEEbiography}[{\includegraphics[width=1in,height=1.25in, clip,keepaspectratio]{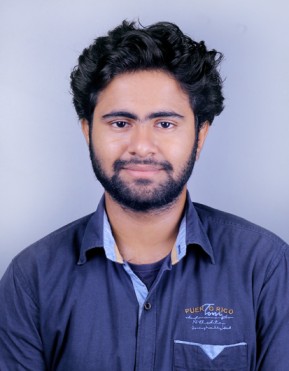}}]{Arkaprabha Basu} received his Bachelor’s and Master’s of Science degrees in Computer Science from Calcutta University in 2018 and Pondicherry University in 2020, respectively. He completed his M.Tech at the University of Hyderabad in 2023. His research interests include Large Language Models, Multimodal Deep Learning, and Computer Vision. He has worked as a Lead Data Scientist at HappyMonk.AI, developing real-time computer vision models for surveillance footage. He is currently a Ph.D. student at the Intelligent Distributed Computing and Systems (INDUCE) Lab, Department of Computer Science and Software Engineering of the United Arab Emirates University.
\end{IEEEbiography}
\begin{IEEEbiography}[{\includegraphics[width=1in,height=1.25in, clip,keepaspectratio]{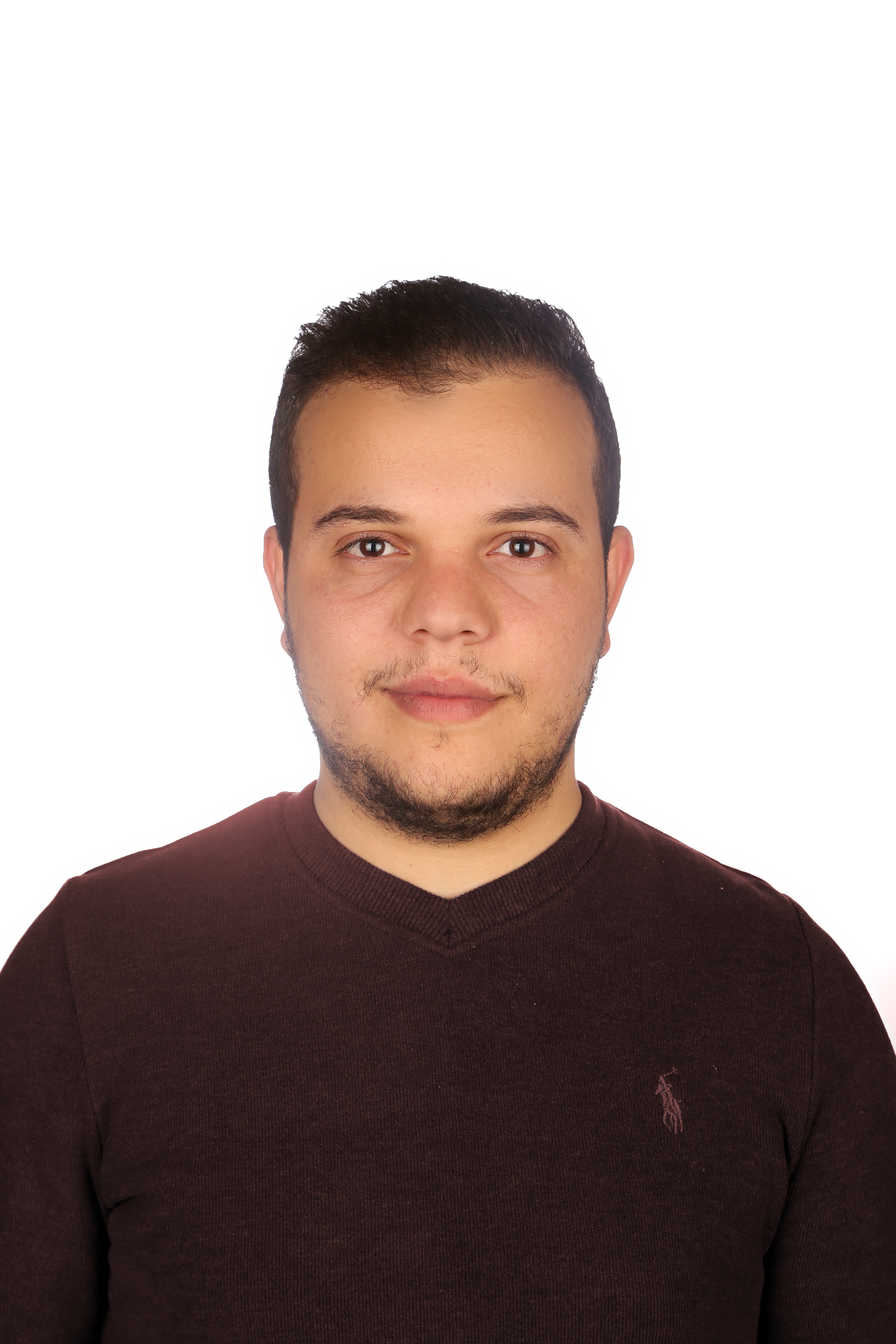}}]{Aymen Dia Eddine Berini} is a postdoctoral researcher at the College of Information Technology, UAE University. He received his Ph.D. in Computer Science from Guelma University, Algeria, in 2023, following his B.Sc. and M.Sc. degrees, both in Computer Science, from the University of Djelfa, Algeria, in 2015 and 2017, respectively. Dr. Berini’s research focuses on critical areas of cybersecurity, with an emphasis on AI security, network security, the Internet of Drones, UAV-assisted terrestrial networks, blockchain, and applied cryptography.
\end{IEEEbiography}
\begin{IEEEbiography}[{\includegraphics[width=1in,height=1.25in, clip,keepaspectratio]{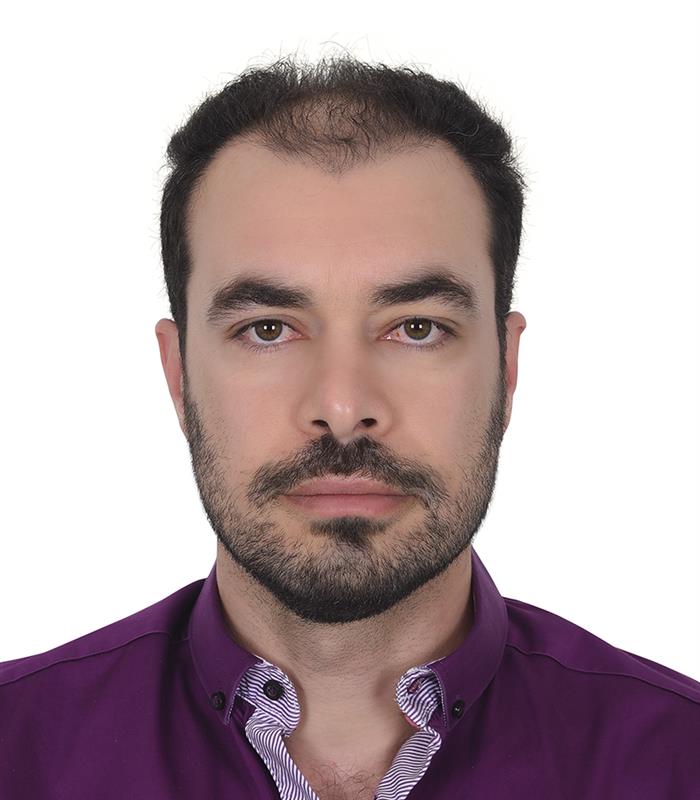}}]{Mohammad Naouss} received the Ph.D. degree in Embedded Electronics from the IMS Laboratory, University of Bordeaux, France, in 2016. He is currently an assistant professor at the College of Information Technology, United Arab Emirates University (UAEU), Al Ain, UAE. His research focuses on computing systems for space and mobility applications, including autonomous systems. His interests include aging modeling of electronic devices—particularly FPGAs—and System-on-Programmable-Chip (SoPC) architectures.
\end{IEEEbiography}
\end{document}